\documentclass[10pt,twocolumn,letterpaper]{article}

\usepackage[pagenumbers]{cvpr} % To force page numbers, e.g. for an arXiv version
\usepackage{xcolor} 
\usepackage[most]{tcolorbox}
\usepackage{colortbl}
\usepackage{bm}
\usepackage{algorithm}
\usepackage{algpseudocode}
\usepackage{stmaryrd}

\newcommand{\cellorg}{\cellcolor{orange!25}}
\newcommand{\cellyel}{\cellcolor{yellow!28}}
\newcommand{\cellred}{\cellcolor{red!25}}
\newcommand{\cellgra}{\cellcolor{gray!25}}

\newcommand{\tightpara}[1]{\vspace{-2.5mm}\paragraph{#1}}

\definecolor{cvprblue}{rgb}{0.21,0.49,0.74}
\usepackage[pagebackref,breaklinks,colorlinks,allcolors=cvprblue]{hyperref}

\def\paperID{33981} % *** Enter the Paper ID here 17741
\def\confName{CVPR}
\def\confYear{2026}

\title{FILTR: Extracting Topological Features from Pretrained 3D Models}

\author{Louis Martinez
\hspace{25pt} 
Maks Ovsjanikov\\
LIX, École Polytechnique, IP Paris\\
{\tt\small louis.martinez@lix.polytechnique.fr}\\
{\tt\small \url{https://filtr-topology.github.io/}}
}
\begin{document}
% \maketitle

% \maketitle
\twocolumn[{
  \maketitle 
  \begin{center}
    \captionsetup{type=figure}
    \centering 
    \includegraphics[width=1.0\textwidth]{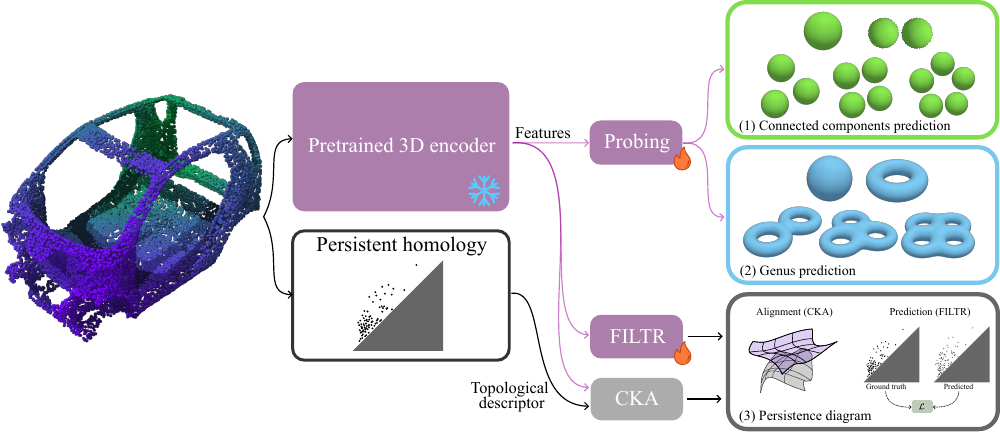}
    \captionof{figure}{We evaluate the topological information implicitly captured by pretrained 3D point-cloud encoders through three distinct tasks. The first two tasks assess whether features produced by modern 3D encoders capture the number of connected components (top) and the genus (middle) of the underlying shapes. We introduce DONUT, a novel benchmark with topological labels, and an adapted probing mechanism. The third task (bottom) evaluates to what extent (i) information contained in persistence diagrams is present in encoder features, and (ii) how it can be extracted. To this end, we propose FILTR (Filtration Transformer), the first model that predicts persistence diagrams directly from pretrained, frozen encoder features, in a feed-forward manner.}
    \label{fig:teaser} 
  \end{center}
}]    
\begin{abstract}

\vspace{-4mm}
Recent advances in pretraining 3D point cloud encoders (e.g., Point-BERT, Point-MAE) have produced powerful models, whose abilities are typically evaluated on geometric or semantic tasks. At the same time, topological descriptors have been shown to provide informative summaries of a shape's multiscale structure. In this paper we pose the question whether topological information can be derived from features produced by 3D encoders. To address this question, we first introduce DONUT, a synthetic benchmark with controlled topological complexity, and propose FILTR (Filtration Transformer), a learnable framework to predict persistence diagrams directly from frozen encoders. FILTR adapts a transformer decoder to treat diagram generation as a set prediction task. Our analysis on DONUT reveals that existing encoders retain only limited global topological signals, yet FILTR successfully leverages information produced by these encoders to approximate persistence diagrams. Our approach enables, for the first time, data-driven extraction of persistence diagrams from raw point clouds through an efficient learnable feed-forward mechanism.
\end{abstract}    
\section{Introduction}
\label{sec:intro}

Recent transformer-based 3D point-cloud encoders, trained on large amounts of data, have demonstrated impressive performance on a wide range of tasks, exhibiting strong generalization capabilities \cite{yu2022pointbert,pang2022pointmae,chen2023pointgpt,wu2024point}. Yet, the structural properties and the full expressive power of the features learned by these encoders remain poorly understood. At the same time, in many real-world applications, ranging from protein structure analysis \cite{xia2014persistent}, material science \cite{obayashi2022persistent}, the study of dynamical systems \cite{maletic2016persistent} and geoscience \cite{janin2025geodynamics} topological invariants have been shown to be  highly informative in characterizing the shape of data. While exploiting topological information for analyzing and processing 3D point clouds has proven beneficial \cite{pathak2025revisiting}, previous approaches typically rely on classical estimation methods, which are decoupled from end-to-end learning-based approaches.

In this paper we ask whether topological information can be extracted directly from the features produced by existing pretrained 3D point cloud encoders. Our motivations are twofold: (1) we aim to shed light on the expressiveness and the potential limitations of current 3D point cloud encoders, by evaluating whether \textit{topological} (rather than semantic or geometric) information can be extracted from their features; (2) informed by this analysis, we seek to enable direct \textit{feed-forward estimation} of topological information. Such an estimator offers significant advantages over classical methods, including computational efficiency and compatibility with other learning-based architectures.

We approach these tasks in several stages. First, we introduce DONUT (Dataset Of maNifold strUcTures), a dataset of meshes carefully labeled according to their number of connected components and genus. We evaluate 3D encoders on this new dataset by probing their features across transformer blocks with trainable decoder modules. We observe relatively modest performance of most existing encoders, suggesting room for improvement on this new task. We then switch focus to estimating persistence diagrams \cite{carlsson2004persistence,carlsson2009topology}, which provide a \textit{multiscale} description of the underlying topology. The structure of persistence diagrams requires using a particular protocol to compare them with representations learned by encoders. We therefore decompose it into two sub-tasks; first we evaluate how well 3D encoder representations \textit{align} with vectorizations of persistence diagrams in a parameter-free way. Then, we introduce FILTR (Filtration Transformer), a framework for \textit{predicting} persistence diagrams from 3D encoder representations, using a trainable transformer architecture. Figure~\ref{fig:teaser} summarizes the tasks and protocols presented in this paper. %For each task, we evaluate the same transformer encoder pretrained with ShapeNet \cite{chang2015shapenet} on three different pretext tasks, namely Masked Point Modeling (Point-BERT \cite{yu2022pointbert}), Masked Auto-Encoding (Point-MAE \cite{pang2022pointmae}) and Auto-regressively generative pre-training (PointGPT \cite{chen2023pointgpt}). 
Remarkably, we find that although 3D encoders have a limited understanding of global topology, it is possible to obtain promising results in predicting \textit{persistence diagrams.} Perhaps even more interestingly, by using a pretrained encoder, FILTR is able to generalize to unseen data distributions. 
Overall, our work aims to both provide a better understanding of topology encoded by 3D feature extractors, and develop fully data-driven feed-forward approaches to extracting topological descriptors.

Our contributions are as follows:

\begin{itemize}
    \item We introduce DONUT, a dataset of synthetic 3D meshes with topological annotations on the number of connected components and genus.
    \item We carry out the first study to quantify how well 3D point-cloud encoders capture topological information through probing and representation alignment.
    \item We introduce FILTR, the first framework for predicting persistence diagrams from 3D point-clouds in a feed-forward manner, and show its generalization capabilities to unseen data distributions.
\end{itemize}

All the code and data to reproduce our experiments are available at \url{https://filtr-topology.github.io/}.
\section{Related work}
\label{sec:related_works}

\paragraph{Self-supervised pretraining on 3D point clouds.}
Self-supervised 3D encoders largely reuse recipes from vision and NLP with minimal conceptual changes: masked language/image modeling becomes masked point modeling \cite{yu2022pointbert,pang2022pointmae,zhang2022pointm2ae,liu2022maskpoint}, contrastive objectives from images transfer to 3D scenes and cross-modal 2D/3D learning \cite{chen2020simclr,he2020moco,grill2020byol,xie2020pointcontrast,afham2022crosspoint}, autoencoding/inpainting pretexts are used via occlusion completion and temporal MAE \cite{wang2021occo,wei2024tmae}, autoregressive language-modeling is translated to point tokens \cite{chen2023pointgpt}, and latent prediction such as JEPA and Data2Vec is directly adapted from their 2D counterpart \cite{saito2025point,knaebel2023point2vec}. Their success, amplified by data scaling in vision and NLP, stems from strong cross-task generalization \cite{he2022mae,chen2020simclr,devlin2019bert,vaswani2017attention}, and promising evidence shows 3D transformer encoders also generalize across downstream tasks, albeit on simpler data distributions than the largest image/text corpora \cite{pang2022pointmae,zhang2022pointm2ae}. However, no prior work quantifies their generalization in topological understanding; to our knowledge, we are the first to explicitly evaluate this aspect, and our FILTR decoder is agnostic to the encoder's pretraining recipe.

\tightpara{Persistence diagram vectorizations.}
Most prior work combining machine learning and topological data analysis (TDA) has focused on converting persistence diagrams into forms that standard ML pipelines can use either by defining kernels on diagrams or by learning vector embeddings of them \cite{reininghaus2015stable,carriere2017sliced,le2018persistencefisher,hofer2017deep,carriere2020perslay}. In our case we take a different path: we predict the persistence diagrams directly from features produced by point-cloud encoders, leveraging the topological signals already present in those features. That is, instead of treating diagrams as a downstream input, we make them the output target of our system. This shift opens up the possibility of measuring how much topology is captured by the learned encoder features and this is the first work to evaluate that. Furthermore, our FILTR decoder does not depend on how the encoder was pretrained; it works regardless of the self-supervised recipe used.

\tightpara{Approximation of persistence diagrams.}
Most work on approximating persistence diagrams falls into two camps. On the algorithmic side, fast approximations with provable error guarantees have been developed for scalar fields \cite{vidal2021fastApproxPD}. On the learning side, recent methods target graphs and build strong inductive biases into the architecture by mirroring steps of the persistent homology computation; they often predict proxy representations rather than full diagrams \cite{de2022ripsnet}, or predict diagrams and then convert them into proxies such as persistence images \cite{yan2022neuralApproxTopFeat,slater2023phseg}. By contrast, we predict persistence diagrams directly from features produced by point-cloud encoders, without hard-coding algorithmic structure into the network. The key idea is to leverage the topological signals already captured—implicitly—by representations learned from large 3D datasets.

\tightpara{Set prediction with transformers.}
Treating a persistence diagram as a set is natural, and transformers built for sets make this feasible in practice. Foundational work establishes permutation-invariant/equivariant modeling for sets and shows how attention can operate directly on them \cite{zaheer2017deepsets,lee2019settransformer}. Set prediction has also been studied directly, where models output an unordered, variable-size collection with appropriate losses \cite{zhang2019dsnp}. In detection, transformers formulate the output as a set and use bipartite matching during training—both in 2D and 3D \cite{carion2020detr,misra2021threedetr}. Recent work further extends attention to multisets, where multiplicities matter, which is directly relevant to diagrams \cite{wang2024multiset}. While these detectors are trained end-to-end from a backbone to a decoder, our pipeline is lighter: FILTR takes features from pretrained point-cloud encoders and predicts the set of diagram points, keeping the encoder fixed.
\section{Do 3D encoders understand topology?}
\label{sec:topology_transformers}

The first key question we pose is whether pretrained 3D point-cloud encoders capture topological information in their learned representations. This task is non-trivial since such encoders are trained either with geometric (e.g., masked auto-encoding) or semantic (e.g., language alignment) losses, which are not directly encouraged to encode topology. To quantify how much topological information is captured by existing 3D point cloud encoders, we identify three tasks (Figure \ref{fig:teaser}): (i) predicting the number of connected components of the shape approximated by the cloud, (ii) predicting its genus, and (iii) degree of alignment between persistence diagrams and encoder features. The first two criteria are evaluated through probing (\cref{ssec:probing}), while (iii) is performed in a parameter-free way with Centered Kernel Alignment (\cref{ssec:alignment}). Probing requires a ground truth dataset annotated with topological labels. To this end, we introduce DONUT (\cref{ssec:donut}), the first benchmark explicitly designed to test topological information present in 3D point cloud encoder features.

\subsection{DONUT: Dataset Of Manifold Structures}
\label{ssec:donut}

\begin{figure}[h]
  \centering
  \includegraphics[width=1.0\linewidth]{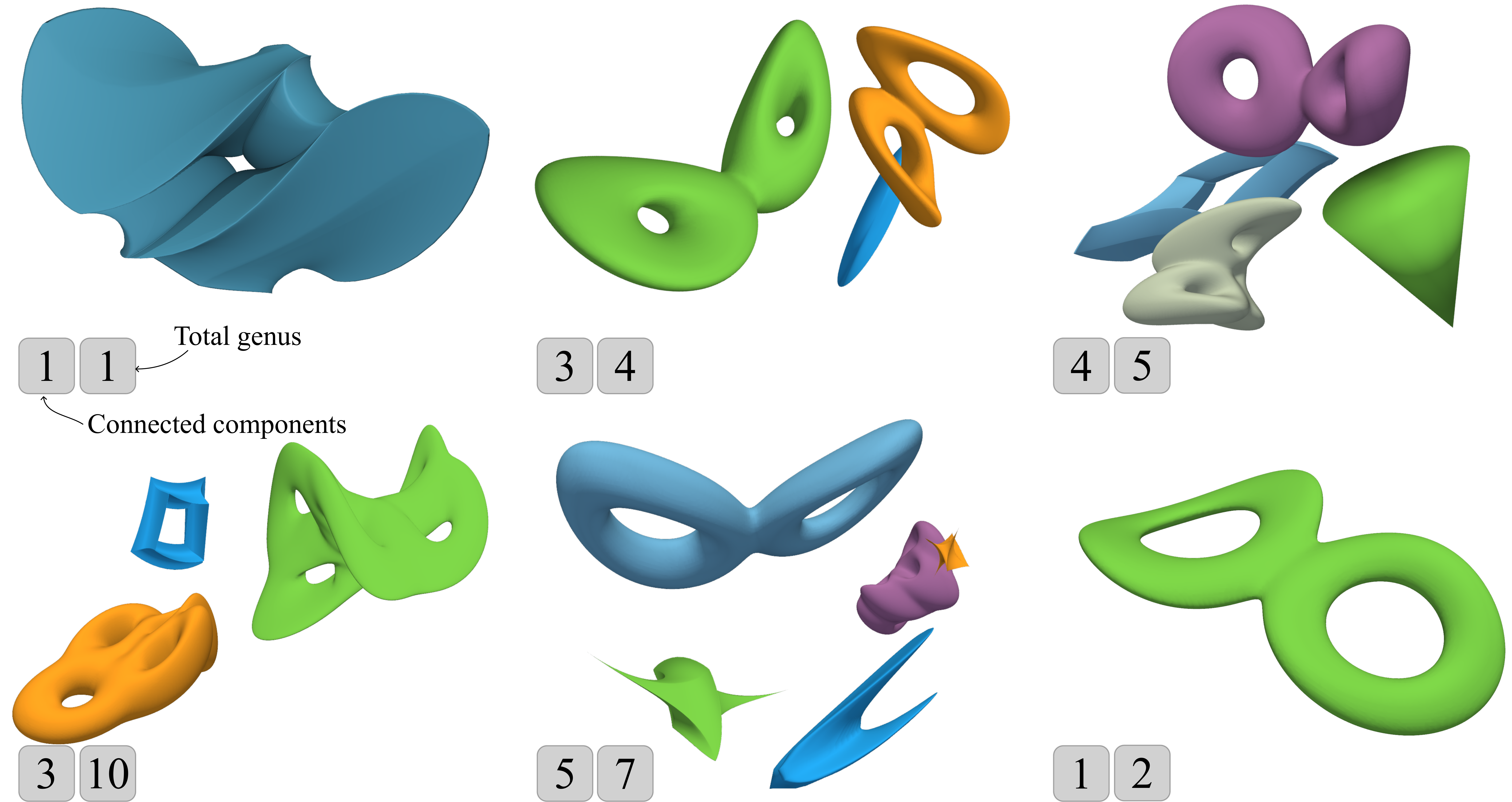}
   \caption{\textbf{Samples from DONUT.} Each object is plotted with its topological labels: number of connected components ($\beta_0$) and the \textit{total} genus ($g$) (the sum of genera across connected components). The dataset is available at \url{https://huggingface.co/datasets/LouisM2001/donut}.}
   \label{fig:donut-samples}
   \vspace{-0.5cm}
\end{figure}

\paragraph{Motivation.}
Most labeled 3D datasets, such as ShapeNet \cite{chang2015shapenet} or ModelNet \cite{wu20153d} are primarily organized by semantic category. While some datasets such as ABC \cite{koch2019abc} and Thingi10K \cite{zhou2016thingi10k} contain topological annotations, unfortunately most shapes in these datasets have a single connected component, and only a fraction of them are topologically richer. Furthermore, we found the reliability of the annotations to be uneven, since the many meshes present in these datasets are either non-manifold, or disconnected. The computation of invariants such as the genus thus becomes unreliable in the presence of such artifacts. Lastly, we note that a \textit{concurrent} effort, EuLearn \cite{fritz2025eulearn}, presents a set of shapes with topological annotations, designed for learning. The focus of that work, however, is on knot-structures, composed of a single connected component, whereas we introduce general surfaces with controlled geometric and topological variability.

Specifically, we propose DONUT, a dataset of manifold structures, with balanced topological annotations (\cref{fig:donut-samples}). Every sample in our dataset is composed of 1 to 6 connected components ($\beta_0$). Each component is a manifold mesh. The total genus $g$ per sample varies from 0 to 10.

\paragraph{Creation.}
The creation of DONUT involves several steps. First we specify the target labels for the whole dataset, to ensure a balanced distribution. The sampling process is further detailed in the supplementary materials. Then, we create a diverse set of parametric shapes (cones, tori and superquadrics), such that their combination satisfies the predefined labels. Finally, we apply a series of geometric transformations to each shape to create variations while preserving their topological properties. 

Synthetic datasets often meet one major pitfall: synthetic shapes are often geometrically too simple, making any downstream task trivial to solve because of undesired shortcuts, such as nearest-neighbor retrieval from the test to the training set. We address this concern by adding as much geometric variety as possible (\cref{fig:donut-samples}), to confound simple retrieval-based approaches. We further apply topology preserving augmentations to preserve labels accuracy.  

% (ii) the shapes within such datasets can fall out of the distribution of data learned by evaluated encoders, making the results meaningless.
% Interestingly, as demonstrated in Figure ... \maks{currently missing}, despite these augmentations, our shapes still lie within the trained distribution of modern 3D point cloud encoders. The full generation process of DONUT is detailed in the Appendix. 

\begin{figure}[t!]
  \centering
  \includegraphics[width=1.0\linewidth]{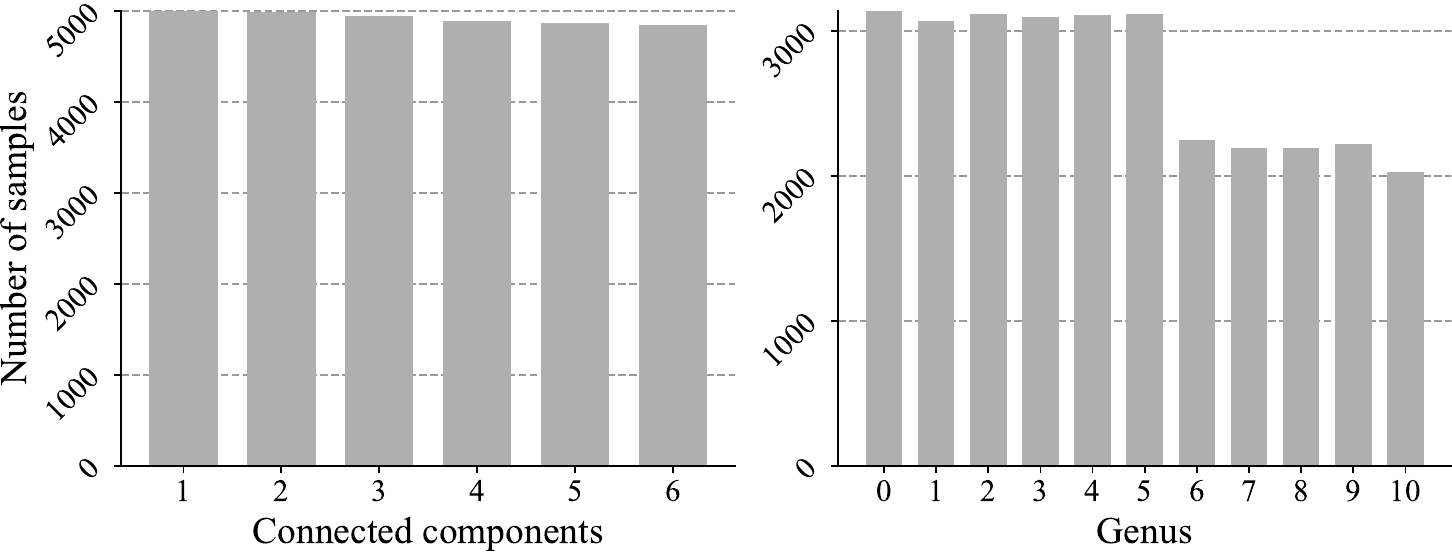}
   \caption{\textbf{Label distribution of DONUT.} We put special care to ensure an even distribution of labels, to avoid biases during training or testing.}
   \label{fig:donut-distrib}
\end{figure}

Overall, DONUT consists of 29,517 objects. Figure \ref{fig:donut-distrib} shows the distribution of the two topological labels. We can see that the dataset is well-balanced across all values of $\beta_0$ and $g$.

\begin{table}[t!]
\centering
\begin{tabular}{l|cc}
\toprule
Model & \#Components & Genus \\
\midrule
\multicolumn{1}{l}{} & \multicolumn{2}{c}{\textit{Probed pretrained models}} \\
\midrule
PointGPT \cite{chen2023pointgpt} & $43.8_{(12)}$ & $22.5_{(6)}$ \\
PCP-MAE \cite{zhang2024pcp} & \cellyel$51.4_{(8)}$ & \cellorg$24.8_{(8)}$ \\
Point-MAE \cite{pang2022pointmae} & $50.0_{(12)}$ & \cellyel$23.1_{(11)}$ \\
Point-BERT (Patch) \cite{yu2022pointbert} & \cellorg$51.5_{(6)}$ & $22.8_{(3)}$ \\
Point-BERT (CLS) \cite{yu2022pointbert} & \cellred$\bm{57.2_{(10)}}$ & \cellred$\bm{25.9_{(7)}}$ \\
\midrule
\multicolumn{1}{l}{} & \multicolumn{2}{c}{\textit{Models trained end-to-end}} \\
\midrule
PointNet \cite{Qi2017PointNet}  & 53.2 & 20.4 \\
PointNet++ \cite{Qi2017PointNetPlusPlus} & 75.7 & \cellorg 51.0 \\
DGCNN \cite{Wang2019DGCNN} & \cellorg 80.8 & 43.5 \\
RepSurf \cite{Ran2022RepSurf} & \cellred \bf 83.3 & \cellred \bf 57.7 \\
\bottomrule
\end{tabular}
\caption{\textbf{Accuracy on DONUT.} For pretrained encoders, we report the best probing accuracy across all transformer layers, with the index of the corresponding layer shown in subscript. For Point-BERT, we probe both the CLS token and the pooled patch tokens. Baseline models (bottom block) are trained end-to-end from scratch on DONUT. Full training details are provided in the Appendix.}
\label{tab:donut-baselines}
\vspace{-0.2cm}
\end{table}

\subsection{Encoders probing}
\label{ssec:probing}

\begin{figure}[t!]
  \centering
  \includegraphics[width=1.0\linewidth]{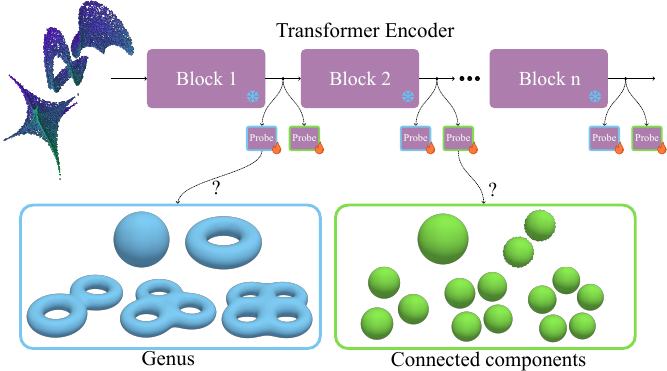}
   \caption{\textbf{Encoder Probing Pipeline.} We probe the features of each (frozen) transformer block on DONUT to predict the number of connected components and the genus.}
   \label{fig:probing-overview}
   \vspace{-0.5cm}
\end{figure}

\paragraph{Problem statement.}
Beyond training models from scratch, we also aim to understand whether topological signal is present in features extracted by modern pretrained point-based encoders. We focus specifically on transformer-based models, as they form the backbone of virtually all state-of-the-art approaches. 

\tightpara{Experimental setup.}
We evaluate four recent 3D point-cloud encoders: Point-BERT \cite{yu2022pointbert}, Point-MAE \cite{pang2022pointmae}, PointGPT \cite{chen2023pointgpt} and PCP-MAE \cite{zhang2024pcp}. All these encoders are pretrained on ShapeNet \cite{chang2015shapenet} on reconstruction tasks. While recent encoders pretrained on latent prediction \cite{saito2025point,knaebel2023point2vec} have demonstrated similar performance on downstream tasks, we theoretically motivate the use of reconstruction-based encoders in the Appendix. We use the weights provided by the authors. While the first three encoders are seminal works in point-cloud pretraining, PCP-MAE is a more recent approach, currently considered state-of-the-art for reconstruction-based encoders. 

We consider two types of features: (i) the CLS token (only for Point-BERT since it is the only encoder with a CLS token), and (ii) the max-pooled patch tokens. 
We probe \textit{each layer} of the transformer architecture to evaluate how the presence of topological information evolves across transformer blocks in pretrained 3D point-cloud encoders. Specifically we probe the output features of each block on DONUT point clouds, sampled with 1024 points, with two separate linear layers: one to predict the number of connected components, the other for the genus, as shown in Figure \ref{fig:probing-overview}. Probing layers are trained with cross-entropy loss. We perform 5-fold cross-validation and report average accuracies. We emphasize that predicting genus and $\beta_0$ from point clouds is non-trivial, especially given the geometric variety in DONUT. These labels are global and invariant under continuous deformation, making them fundamentally harder to infer from local features alone.

\begin{figure}[t]
  \centering
  \includegraphics[width=1.0\linewidth]{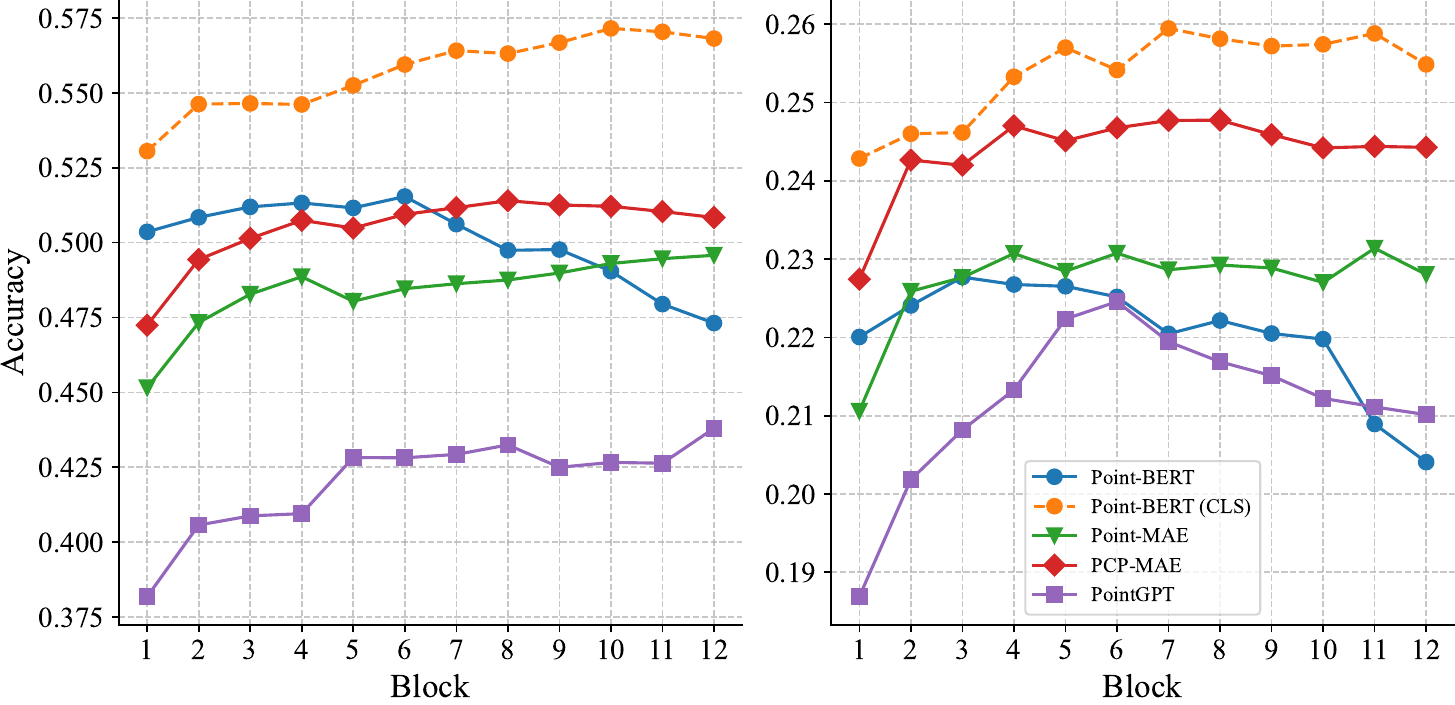}
   \caption{\textbf{Layer-wise performance on DONUT.} We report probing accuracies for different encoders, on number of connected components \textit{(left)} and genus \textit{(right)}. Unlike the other encoders, Point-BERT is pretrained with a CLS token, which we also probe (dashed line).}
   \label{fig:probing-results}
   \vspace{-0.5cm}
\end{figure}

\tightpara{Results.} Table \ref{tab:donut-baselines} summarizes our main results. Overall, probing pretrained encoders yields low accuracies on both tasks. Their performance is only marginally better than PointNet trained from scratch, showing that current 3D pretraining strategies do not strongly encode topological information. Among all pretrained models, Point-BERT using the CLS token achieves the highest probing accuracy, outperforming its patch-token variant as well as all MAE-based methods. Despite their different pretraining objectives, Point-BERT (Patch), Point-MAE, and PCP-MAE obtain similar results, suggesting that masked reconstruction alone does not facilitate topology-aware representations. PointGPT performs the worst among pretrained encoders, indicating that generative modeling of point sequences may not preserve global structural cues. Interestingly, apart from Point-BERT (Patch), the best probing accuracy is obtained by deeper blocks of the encoders. Figure \ref{fig:probing-results}, which shows probing performance across transformer blocks, confirms these observations. Point-BERT (Patch) aside, accuracy increases in deeper blocks.

In contrast, end-to-end baselines perform substantially better than the probed layers, although their performance on genus prediction remains limited. RepSurf \cite{Ran2022RepSurf} achieves the highest accuracy on both tasks. Its strong results likely stem from its explicit use of surface-based features, which appear beneficial for capturing structural properties of 3D shapes.

\subsection{Features alignment with persistence diagrams}
\label{ssec:alignment}

While probing provides an estimate of the understanding of the \textit{global} structure of shapes, encoder features might also carry information about \textit{fine-grained} topological structures, at different scales. In parallel, persistence diagrams are specifically tailored to provide a multiscale description of the structure of point-clouds. In addition, numerous methods have been proposed to vectorize these descriptors \cite{giusti2015clique, bubenik2015persistence}. We therefore use Centered Kernel Alignment (CKA) to quantify the similarity between encoder features and these vectorizations. This provides a solid proxy to quantify the multiscale information captured by encoders.

CKA measures the similarity between two sets of representations. It has been frequently adopted \cite{davari2023reliability,pmlrv97kornblith19a} to compare learned features from different models or layers within a model. We further refer the reader to Kornblith et al. \cite{pmlrv97kornblith19a} for a detailed explanation of CKA.

\tightpara{Experimental setup.} We compare encoder features with two types of vectorizations: Analytic and learned. Analytic vectorizations are fixed, closed-form mappings from a persistence diagram to a feature vector. Learned vectorizations, such as ATOL~\cite{royer2021atol}, are trained in an unsupervised way to map diagrams to vectors from the empirical distribution of diagrams. Here, we use only $\mathcal{H}_1$ persistence diagrams computed from the $\alpha$-filtration \cite{Edelsbrunner1994AlphaShapes} of point clouds with 1024 points. Although Vietoris-Rips filtration is popular in TDA, its computational cost \cite{Zhang2020veitoris} limits its use to even small point clouds ($\sim 10^3$ points). For each encoder and transformer block, we then compute CKA between these vectors and the corresponding features. We report results on 23\,579 random samples from DONUT.  

\begin{figure}[t]
  \centering
  \includegraphics[width=1.0\linewidth]{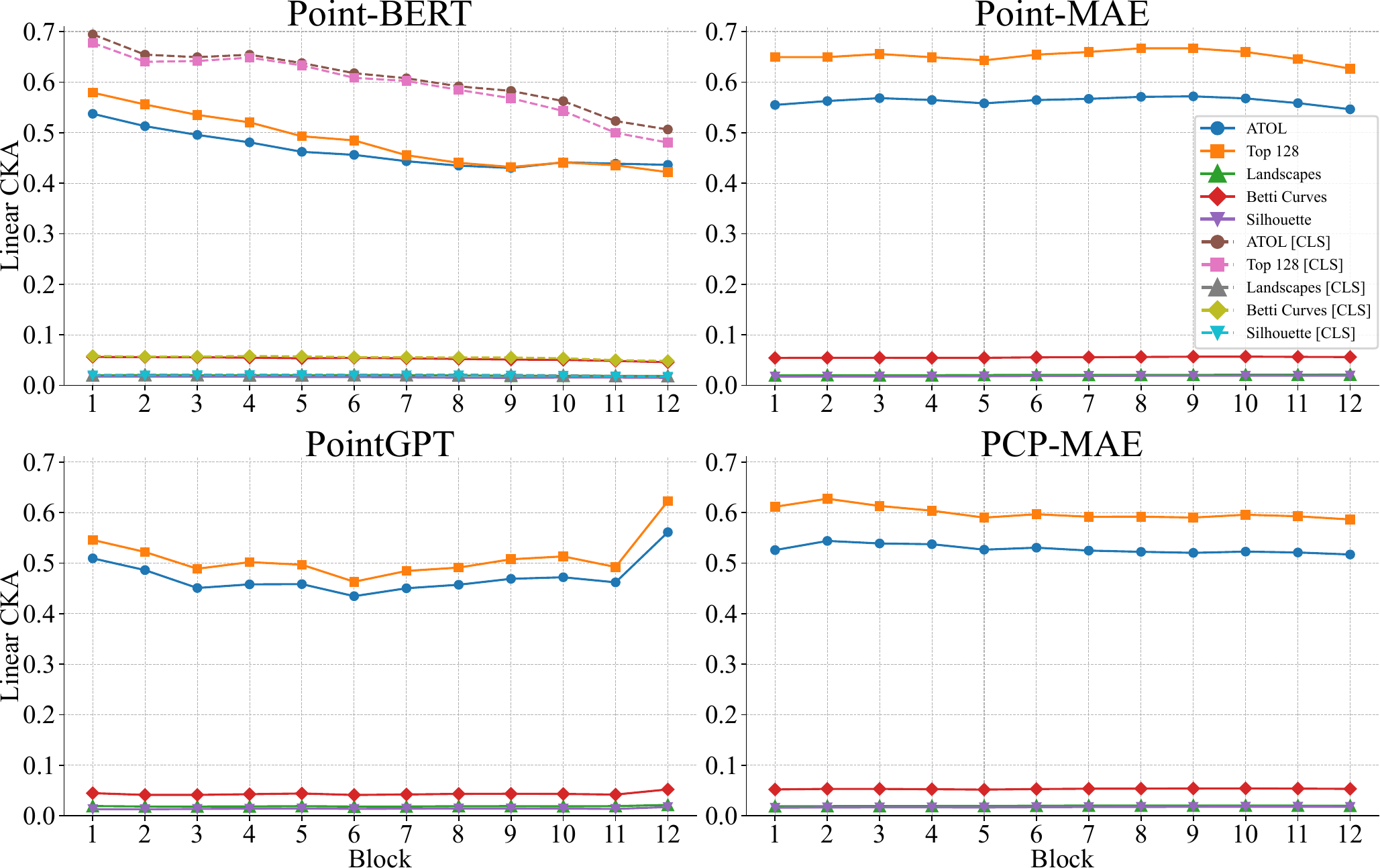}
   \caption{\textbf{CKA results on DONUT.} We report linear CKA scores between encoder features and persistence diagram vectorizations for different encoders and vectorization methods. The higher the score, the stronger the alignment. [CLS] refers to the CLS token of Point-BERT.}
   \label{fig:cka-results}
   \vspace{-0.5cm}
\end{figure}

\tightpara{Results.} Figure \ref{fig:cka-results} shows that MAE-based models, especially Point-MAE, align consistently across layers with vectorized persistence diagrams. Unlike the probing results in Figure \ref{fig:probing-results}, we see no substantial gain in similarity in deeper blocks. We hypothesize that this is due to how point clouds are processed: they are first patchified and embedded, so each patch starts with mainly local information. As attention layers mix these patches, some global structure appears, but the original local signals are preserved. We provide in the Appendix results for denser point-clouds (2048 points).
\section{FILTR}
\label{sec:filtr_method}

\begin{figure}[h]
  \centering
  \includegraphics[width=1.0\linewidth]{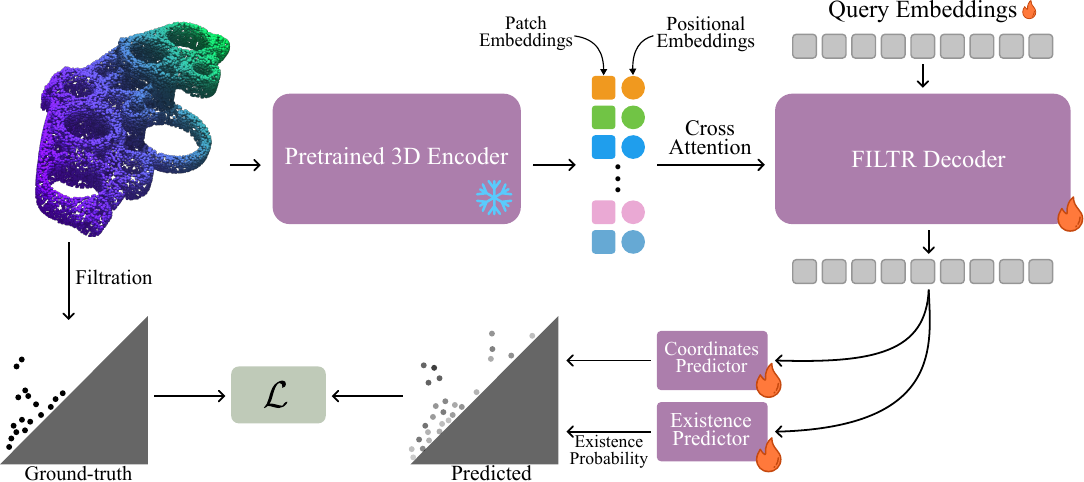}
   \caption{\textbf{FILTR Pipeline.} A frozen 3D point-cloud encoder produces features and positional encodings. These condition the decoder through cross-attention. The decoder processes a fixed set of learned queries to predict persistence pairs and their existence probabilities (shown as gray intensities). Training uses a set-prediction loss to match predicted and ground-truth pairs.}
   \label{fig:filtr-overview} 
\end{figure}

Section \ref{sec:topology_transformers} revealed that while being limited, pretrained 3D encoders capture some multiscale topological information in their learned representations. This motivates us to leverage these encoders as feature extractors to predict persistence diagrams. In this section, we introduce FILTR, a novel framework designed for this task. We first formalize the problem of predicting persistence diagrams from point clouds. Then, we present how we derive FILTR from DETR \cite{carion2020detr}.

\subsection{Problem definition}
\label{ssec:problem_definition}

Given a point cloud $X = \{x_i\} \subset \mathbb{R}^3$, we aim to predict a persistence diagram $D_q(X)$. Formally speaking, a persistence diagram is a multiset; however, in practice, it is rare to have identical persistence pairs in diagrams computed from point clouds. Thus, we treat $D_q(X)$ as a set of pairs $\{(b_i, d_i)\}_{i=1}^M$, where $M$ is the number of topological features in dimension $q$. Short-lived pairs near the diagonal $\Delta$ often reflect spurious signal (topological noise). Two common strategies remove this noise: (i) statistical procedures, often relying on bootstrapping \cite{Chazal2015Subsampling,Kwitt2015KernelTDA,fasy2014confidence}; (ii) heuristics that keep only the most persistent features (e.g. top-$k$ or a persistence quantile) \cite{carriere2020perslay,Wu2024persistence,reininghaus2015stable,carriere2017sliced}. We adopt (ii) via a fixed persistence quantile, since statistical procedures aim at recovering the true homology of the whole point cloud. Therefore, they prune persistence diagrams more aggressively than a quantile-based approach, losing track of local topological cues. Similarly to Section \ref{ssec:alignment}, we opt for the $\alpha$-filtration for computational efficiency. However, our framework is agnostic to the choice of filtration; we provide additional results using the Vietoris-Rips filtration in the Appendix. 

\subsection{Adapting DETR architecture}
\label{ssec:adapt_detr}

\begin{table}[t]
\centering
\resizebox{\linewidth}{!}{
\begin{tabular}{l|cc}
\toprule
Concept & \textbf{DETR} & \textbf{FILTR} \\
\midrule
Input & Image & Point cloud \\
Target & Bounding box & Persistence pair \\
Training & End-to-end & Pretrained encoder\\
Positional encoding & 2D sinusoidal & 3D patch centers \\
Output constraints & Box in $[0,1]^4$ & $(b,d)$ with $d>b$ \\
Null class & ``No-object'' & ``No-pair'' \\
Matching cost & Class + box loss & Existence + pair regression \\
Regularization & None & Diagonal regularizer \\
\bottomrule
\end{tabular}
}
\caption{\textbf{Core DETR-FILTR analogies.}}
\label{tab:detr_filtr_diff}
\vspace{-0.5cm}
\end{table}

DETR \cite{carion2020detr} frames object detection as a set prediction problem. It is therefore a natural choice for persistence diagrams predictions. Table \ref{tab:detr_filtr_diff} summarizes the adaptations made to DETR.

\tightpara{Features extraction.} A point cloud $X \in \mathbb{R}^{p\times 3}$ is encoded by a frozen 3D backbone into patch features $F = \{f_i\}_{i=1}^n$. Each feature, together with its 3D positional encoding, is projected to the decoder dimension $d_{\text{dec}}$.

\tightpara{Decoder.} The decoder receives $N$ learned query embeddings, where $N$ exceeds the maximum diagram size. As in DETR, they interact with encoder features through cross-attention. The final decoder states feed two MLP heads: one mapping each query to persistence logits $(\hat{p}^{(1)}_i,\hat{p}^{(2)}_i)$, the other producing an existence logit $\hat{l}_i$. Persistence pairs are obtained via $\hat{b}_i=\sigma (\hat{p}_i^{(1)})$, $\hat{d}_i= \hat{b}_i + \text{softplus} (\hat{p}_i^{(2)})$, which enforces the birth–death ordering. Existence probabilities $\sigma(\hat{l}_i)$ indicate whether a query corresponds to a genuine topological feature or a no-pair slot.

\subsection{Set prediction loss} 
\label{ssec:set-pred-loss}
A principled option is to train with a 2-Wasserstein loss between predicted and ground-truth diagrams, letting unmatched predictions flow to the diagonal. While this has been successful in settings with small diagrams and strong architectural priors (e.g., Yan et al. on graphs~\cite{yan2022neuralApproxTopFeat}), we empirically found it unreliable for point-clouds where diagrams frequently exceed $10^2$ pairs.

FILTR therefore adopts a set-prediction objective: (i) Hungarian matching with a coordinate regression term; (ii) a binary existence loss to decide on/off-diagonal status; and (iii) a diagonal regularizer that pushes non-matched predictions toward the diagonal, making thresholding largely optional. The full loss is: 

\begin{equation}
    \mathcal{L} = \mu_\text{recon} \mathcal{L}_\text{recon} + \mu_\text{exist} \mathcal{L}_\text{exist} + \mu_\text{diag} \mathcal{L}_\text{diag}.
\end{equation}

\tightpara{Pairs matching and reconstruction loss.}
FILTR outputs $N$ \textit{unordered} persistence pairs $\{\hat{y}_i\}_{i=1}^N$. We compute an assignment $\pi^*:\{1,\ldots,M\}\rightarrow \{1,\ldots,N\}$ between predicted and ground-truth pairs $\{y_j\}_{j=1}^M$ using the Hungarian algorithm assignment. $\pi^*$ satisfies:

\begin{equation} 
    \pi^* = \arg\min_{\pi} \sum_{i=1}^M \mathcal{L}_\text{match}\big(\hat{y}_{\pi(i)},y_i\big)
\end{equation}

\vspace{-6mm}

\begin{equation}
    \mathcal{L}_\text{match}\big(\hat{y}_i,y_j\big) = \lambda_\text{reg}\|\hat{y}_i-y_j\|_2^2 + \lambda_\text{exist} (1-\sigma(\hat{l}_i))
\end{equation}

$\mathcal{L}_\text{match}$ takes into account the distance between the predicted and ground-truth pairs, as well as the existence score of the predicted pair. If the decoder predicts a pair with a small existence probability, the latter should be more penalized in the matching cost.

Once the optimal assignment $\pi^*$ is found, we define the reconstruction loss as the mean squared error (MSE) over matched pairs:

\vspace{-4mm}

\begin{equation}
    \mathcal{L}_\text{recon} = \frac{1}{M} \sum_{i=1}^M \|\hat{y}_{\pi^*(i)}-y_i\|_2^2.
\end{equation}

\vspace{-4mm}

\tightpara{Existence loss.} 
Existence logits are supervised through a binary cross-entropy loss. For each matched pair, the target existence label is 1, while for unmatched predicted pairs, it is 0. The existence loss is defined as:
\begin{equation}
    \mathcal{M} = \{\pi^*(i)\mid i=1,\ldots,M\}, \qquad \bar{\mathcal{M}} = \{1,\ldots,N\}\setminus\mathcal{M}.
\end{equation}

\vspace{-5mm}

\begin{equation}
    \mathcal{L}_\text{exist} = -\frac{1}{N} \left( \sum_{i=1}^M \log \sigma(\hat{l}_{\pi^*(i)}) + \sum_{j\in\bar{\mathcal{M}}} \log (1 - \sigma(\hat{l}_j)) \right).
\end{equation}

\tightpara{Diagonal loss.}
At inference time, persistence diagrams are obtained by thresholding existence probabilities, typically at 0.5. This can be unstable, leading to poor diagram approximation under standard distances. To mitigate this, we force unmatched predictions to lie near the diagonal, so that their contribution to the diagram distance is negligible and thresholding becomes optional. For each unmatched predicted pair $\hat{y}_i = (\hat{b}_i, \hat{d}_i)$, we penalize its squared distance to the diagonal. The diagonal loss is:
\vspace{-3mm}

\begin{equation}
    \mathcal{L}_\text{diag} = \frac{1}{|\bar{\mathcal{M}}|} \sum_{j\in\bar{\mathcal{M}}} (\hat{d}_j - \hat{b}_j)^2.
\end{equation}

\vspace{-4mm}

\subsection{Experiments}
\label{sec:experiments}

\begin{table*}[h]
\centering
\resizebox{\textwidth}{!}{
\begin{tabular}{l|ccc|ccc|ccc}
\toprule
& \multicolumn{3}{c|}{\textbf{DONUT}} & \multicolumn{3}{c|}{\textbf{ModelNet40}} & \multicolumn{3}{c}{\textbf{ABC}} \\
% \cmidrule{lr}{2-4} \cmidrule{lr}{5-7}
Features extractor & $W_{2\,(\times 10^{-2})}$ & $d_{B\,(\times 10^{-3})} $ & PIE & $W_{2\,(\times 10^{-2})}$ & $d_{B\,(\times 10^{-3})} $ & PIE & $W_{2\,(\times 10^{-2})}$ & $d_{B\,(\times 10^{-3})} $ & PIE \\
\midrule
\multicolumn{1}{l}{} & \multicolumn{9}{c}{\textit{Pretrained-frozen encoders}} \\
\midrule
PointGPT$_{\text{L}}$   & 17.86               & 10.03               & \cellorg 1.192      & \cellred \bf{39.80} & \cellorg 12.63      & 10.30               & \cellred \bf{40.19} & \cellred \bf{31.73} & \cellyel 3.799 \\
PointGPT$_{\text{C}}$   & 17.59               & 10.08               & 1.289               & 48.89               & 12.99               & 10.22               & \cellyel 40.47      & 32.14               & 3.806 \\
\midrule
PCP-MAE$_{\text{L}}$    & \cellyel 17.18      & 9.950               & 1.333               & 46.93               & 13.02               & \cellyel 10.18      & 45.09               & 32.13               & 4.008 \\
PCP-MAE$_{\text{C}}$    & 18.73               & 9.933               & 2.654               & 45.92               & \cellyel 12.87      & \cellorg 9.968      & 46.83               & 32.05               & \cellorg 3.793 \\
\midrule
Point-MAE$_{\text{L}}$  & 17.24               & \cellyel 9.917      & \cellred \bf{1.107} & 45.68               & 13.11               & 10.29               & 46.15               & 32.05               & 3.868 \\
Point-MAE$_{\text{C}}$  & \cellred \bf{16.02} & 9.838               & \cellyel 1.214      & 47.26               & 13.80               & 11.93               & 47.99               & 32.72               & 4.393 \\
\midrule
Point-BERT$_{\text{L}}$ & \cellorg 16.18      & \cellred \bf{9.901} & 1.371               & \cellyel 43.04      & \cellred \bf{12.54} & 10.65               & 43.37               & \cellorg 31.87      & 3.867 \\
Point-BERT$_{\text{C}}$ & 17.33               & \cellorg 9.918      & 1.584               & \cellorg 41.92      & 13.09               & \cellred \bf{9.900} & \cellorg 43.23      & \cellyel 32.04      & \cellred \bf{3.780} \\
\midrule
\multicolumn{1}{l}{} & \multicolumn{9}{c}{\textit{Encoders trained end-to-end}} \\
\midrule
PointNet    & 24.85 & 10.61 & 1.442 & 51.32 & 12.72 & 9.899 & 57.44 & 32.26 & 3.849 \\
PointNet++  & \cellgra 39.37 & 13.64 & \cellgra 10.39 & 78.86 & 12.98 & \cellgra 314.0 & \cellgra 95.39 & 32.13 & \cellgra 39.13 \\
DGCNN       & 16.62 & 10.15 & 1.213 & 43.27 & 12.51 & 9.773 & 45.21 & 31.70 & 3.760 \\
\bottomrule
\end{tabular}
}
\caption{\textbf{Reconstruction results of FILTR.} All the models are trained on DONUT, and evaluated on: a held-out test set from DONUT, ModelNet40 test set, a subset of ABC. We use the same configuration for all pretrained backbones, and report results obtained by training FILTR with either the features of the last transformer block (L), or a combination of the features from all transformer blocks (C) (see Fig. \ref{fig:filtr-baseline} \textit{(left)}). We highlight PointNet++ for its remarkably higher reconstruction errors compared to other architectures. We discuss this point and provide training details in the Appendix.}
\label{tab:filtr-recon}
\vspace{-0.5cm}
\end{table*}

\tightpara{Data and preprocessing.}
FILTR is trained on 23\,579 meshes from DONUT. Each mesh is sampled with 1024 points, and persistence diagrams are computed from these point clouds. We keep the same point clouds across all experiments. Again, we compute $\mathcal{H}_1$ persistence diagrams from the $\alpha$-filtration of point clouds. We keep the 10\% most persistent pairs in each diagram to discard noise. This threshold offers a good trade-off between noise reduction and information preservation. All persistence diagrams are scaled dataset-wise, so that the maximum birth and death values are in the range $[0, 1]$. We evaluate FILTR on a held-out test set of 5\,938 samples from DONUT, the test sets of ModelNet40 \cite{wu20153d}, as well as a subset of 3K samples from ABC \cite{koch2019abc}.

\tightpara{Evaluation metrics.}
We reuse the same metrics as Yan et al. \cite{yan2022neuralApproxTopFeat}: (i) the 2-Wasserstein distance $W_2$ between predicted and ground-truth diagrams and (ii) the Persistence Image Error (PIE). The PIE is the total square error between the ground truth and predicted persistence images. We also report (iii) the bottleneck distance $d_B$. Both $W_2$ and $d_B$ are relevant to consider since they capture different aspects of the prediction quality. $d_B$ is only sensitive to the worst predicted point, while $W_2$ reflects the overall quality of the prediction. Note that the PIE is always computed on persistence diagrams with thresholded pairs. Indeed, persistence images are obtained by placing a smooth kernel at each point of the persistence diagram and integrating the resulting function over a fixed grid. This construction evaluates the kernel at every point, so even pairs lying very close to the diagonal contribute to the image unless they are explicitly removed.

\begin{figure}[t]
  \centering
  \includegraphics[width=1.0\linewidth]{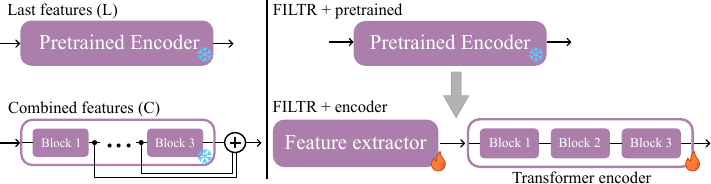}
   \caption{\textit{(left)}  The (L) variant of FILTR (top) only uses the output features of the encoder while the (C) variant sums the features of all intermediate blocks. \textit{(right)} The pretrained frozen encoder is replaced by a feature extractor and a lightweight transformer encoder, both trainable.}
   \label{fig:filtr-baseline} 
   \vspace{-0.5cm}
\end{figure}

\tightpara{Choice of input features.}
We know from Sections \ref{ssec:probing} and \ref{ssec:alignment} that it is unclear from which block of the encoders topological information can be retrieved. Deeper blocks show better \textit{global} understanding, while \textit{fine-grained} information is more spread out. We therefore train two variants of FILTR; we feed the decoder either with (i) the features from the last encoder block, or (ii) the sum of features from all blocks. Figure \ref{fig:filtr-baseline} (\textit{left}) illustrates both strategies.

\tightpara{Baseline.}
We seek to demonstrate that topology captured by pretrained 3D encoders can be efficiently leveraged to predict persistence diagrams. To this end, we replace the pretrained encoder with a point-wise feature extractor followed by a lightweight transformer encoder. Both modules are trained along with the decoder, as shown in Figure \ref{fig:filtr-baseline} (\textit{right}). We use PointNet, PointNet++, and DGCNN as feature extractors. Results with RepSurf are provided in the Appendix, since it relies on a PointNet++ backbone.

\subsection{Results}
\label{ssec:filtr_results}

We report in the Appendix full training details and computational metrics (\cref{tab:comp-cost}).

\tightpara{Features extractor comparison.} Table \ref{tab:filtr-recon} shows that FILTR with frozen pretrained encoders reaches or surpasses the performance of end-to-end baselines, except for the pathological PointNet++ case discussed in the Appendix. This is notable because the probing results showed that these encoders do not linearly expose topological information (Fig. \ref{fig:probing-results}). Combined with the CKA analysis, the most consistent explanation is that pretrained transformers preserve useful local geometric structure, even if they do not directly encode topology. FILTR can exploit this structure through its non-linear decoder and recover accurate diagrams.

The relative behavior of the pretrained models also changes compared to earlier experiments. PointGPT and Point-BERT—previously weaker—now give some of the strongest results on out-of-distribution datasets. In contrast, Point-MAE and PCP-MAE show sharper degradation under distribution shift, despite performing well on the DONUT test set. This suggests that their features are more tied to the statistics of their pretraining data. We also do not observe a systematic benefit of using last-block features versus block-combined features. Finally, the end-to-end baselines follow broadly the same trends but require substantially more trainable parameters, making FILTR a more efficient solution when strong pretrained encoders are available. However, we notice that the DGCNN baseline slightly outperforms pretrained encoders on ModelNet and ABC for the bottleneck distance and PIE. We further discuss this observation in the Appendix.
\vspace{-3pt}
\tightpara{Metric comparison.} The three metrics reveal different error modes. The drop in $W_2$ when moving from DONUT to ModelNet40 and ABC indicates limited cross-dataset generalization. Yet the bottleneck distance increases sharply only on ABC, pointing to a few severe mismatches rather than a uniform degradation. PIE shows the opposite behavior: its increase is much larger on ModelNet40 than on ABC. Since high-persistence points dominate the persistence image, this implies that FILTR makes more mistakes on the most important features of ModelNet40 shapes, while its errors on ABC are mostly on low-persistence, less informative pairs. Together, these patterns indicate that FILTR preserves the overall structure of diagrams under distribution shift, but the nature of the remaining errors depends strongly on the target dataset.

% \tightpara{Loss ablation.}

% Table \ref{tab:filtr-ablation-loss} shows that $\mathcal{L}_\textbf{exist}$ substantially improves reconstruction, compared to $\mathcal{L}_\textbf{exist}$ alone. As expected, without $\mathcal{L}_\textbf{diag}$, it is essential to threshold non-existing persistence pairs to get good performance. Finally, while introducing $\mathcal{L}_\textbf{diag}$ slightly reduces performance for $W_2$, they remain identical with and without thresholding. We also note that overall, the Bottleneck distance $d_B$ seems to be hardly impacted by these variants.    
\section{Conclusion, limitations and future work}
\label{sec:conclusion}

This work presents the first systematic examination of the topological competence of pretrained 3D point-cloud encoders. Our analysis shows that these models capture only weak global topological information but nonetheless display nontrivial correlations with vectorized persistence diagrams, indicating a degree of local structural awareness. To support rigorous evaluation, we introduced DONUT, a dataset with precise topological annotations.

Building on these observations, we showed that persistence diagrams can be approximated directly from pretrained encoder features, offering a feed-forward alternative to classical topological pipelines. While this provides both theoretical and practical insights into designing data-driven topological proxies, such approaches remain inherently constrained by the availability and quality of pretrained encoders. Consequently, in domains such as graph learning—where topology is central but strong general-purpose encoders are still lacking—our conclusions do not yet transfer.

A natural extension is to investigate multimodal foundation models. Modalities such as text may encode structural or relational information in ways that differ from 2D and 3D models, potentially revealing alternative pathways for topological reasoning in large pretrained systems. 

\tightpara{Acknowledgements} Parts of this work were supported by the ERC Consolidator Grant 101087347 (VEGA), as well as gifts from Ansys Inc., and Adobe Research.

{
    \small
    \bibliographystyle{ieeenat_fullname}
    \bibliography{main}
}

\clearpage
\setcounter{page}{1}
\maketitlesupplementary

We provide implementation details, including the creation of DONUT (\cref{ssec:donut-creation}) and the architecture of FILTR and baselines (\cref{ssec:filtr-implem}) along with the training procedure. We also present additional experimental results for probing (\cref{ssec:donut-probing-results}) and feature alignment (\cref{ssec:cka-relevance}), as well as additional experiments to further motivate design choices for FILTR (\cref{ssec:add-filtr}). Finally, we include qualitative results on persistence diagram prediction. All the code and data to reproduce our experiments are available at \url{https://filtr-topology.github.io/}.

\section{Implementation details}
\label{sec:implem}

\subsection{Creation of DONUT}
\label{ssec:donut-creation}

\begin{figure}[h]
  \centering
  \includegraphics[width=1.0\linewidth]{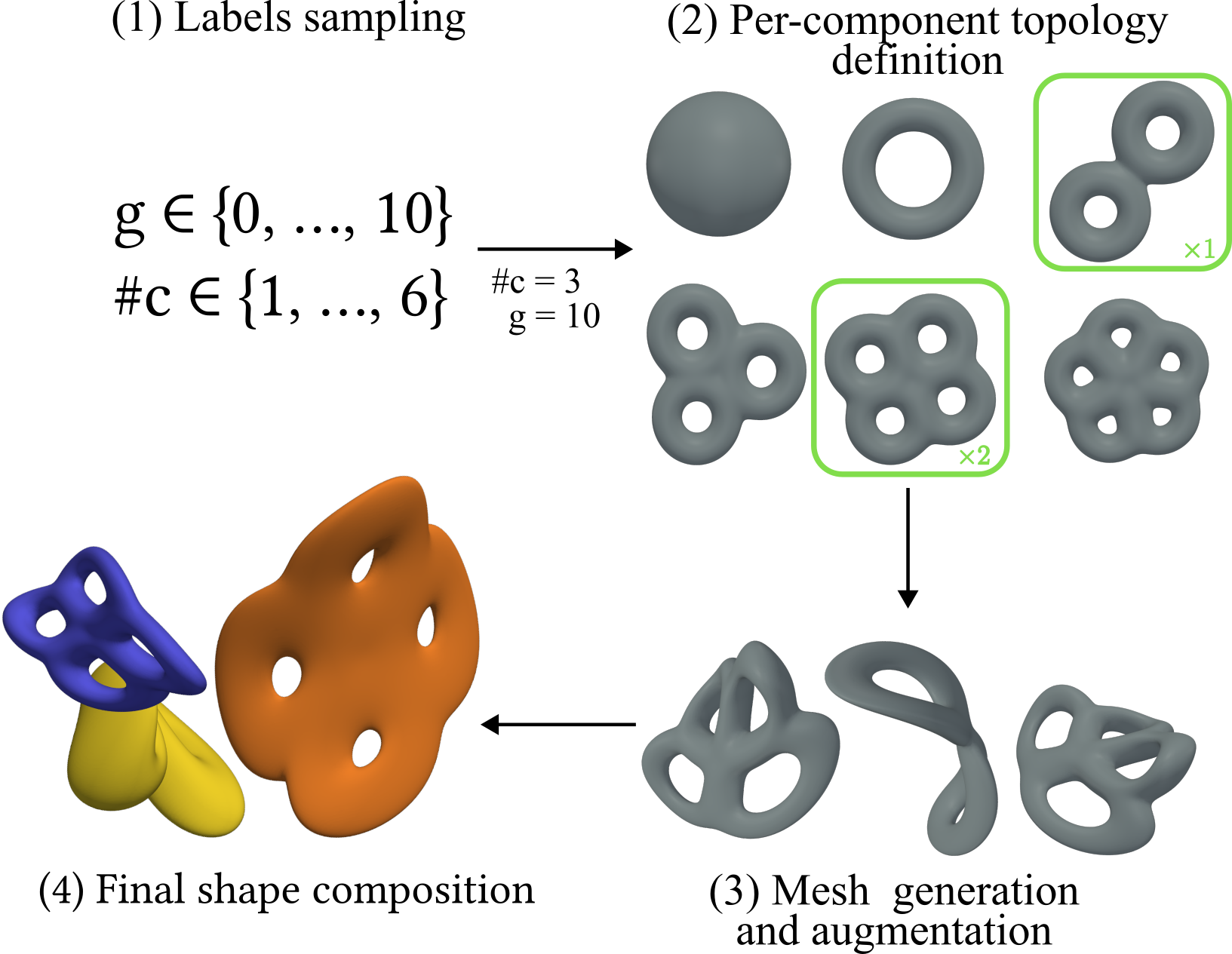}
   \caption{\textbf{DONUT generation pipeline.}  (1) Sample global topological labels (Alg.~\ref{alg:donut-labels}); (2) distribute them across components (Sec.~\ref{ssec:donut-label-sampling}); (3) generate each component mesh (Sec.~\ref{ssec:donut-shape-generation}); (4) apply component-wise augmentations and merge them without overlap to preserve global topology.}
   \label{fig:donut-gen} 
\end{figure}

The primary goal in constructing DONUT is to obtain reliable and balanced topological annotations. The generation pipeline (Fig.~\ref{fig:donut-gen}) therefore first samples valid global labels, then distributes them across components, and finally produces geometrically diverse meshes consistent with the prescribed topology.

\subsubsection{Labels sampling}
\label{ssec:donut-label-sampling}

\begin{table}[h]
\centering
\begin{tabular}{l c}
    \toprule
    \textbf{Hyperparameter} & \textbf{Value} \\
    \midrule
    $g^{\max}$ & 5 \\
    $G^{\max}$ & 10 \\
    $\beta_0^{\min}$ & 1 \\
    $\beta_0^{\max}$ & 6 \\
    $k$ & 1000 \\
    \bottomrule
\end{tabular}
\caption{Hyperparameter values used to create DONUT.}
\label{tab:donut-hyperparams}
\end{table}

Label generation is performed prior to mesh construction and is controlled by a small set of hyperparameters. For each sample, we draw its number of connected components and total genus under the following constraints:
\begin{itemize}
    \item The total genus does not exceed $G^{\max}$.
    \item The genus of each component does not exceed $g^{\max}$.
    \item The number of connected components lies in $\llbracket \beta_0^{\min},\, \beta_0^{\max} \rrbracket$.
    \item The marginal distribution of labels is approximately uniform.
\end{itemize}

Algorithm~\ref{alg:donut-labels} summarizes this sampling. The values of $G^{\max}$, $g^{\max}$, $\beta_0^{\min}$ and $\beta_0^{\max}$ are provided in Table~\ref{tab:donut-hyperparams}.  
After sampling global labels, we assign per-component genera such that they sum exactly to the global genus. This is achieved via a backtracking procedure (Algorithms~\ref{alg:donut-enum-sol}–\ref{alg:donut-backtrack}).

\begin{algorithm}
  \caption{Sampling $(\beta_0, g)$ \\ 
  \textbf{Input:} $g^{\max},\; G^{\max},\; \beta_0^{\min},\; \beta_0^{\max},\; k$} \label{alg:donut-labels}

  \begin{algorithmic}[ ] % you can use [1] to turn on line numbering
    \State let $\mathfrak{B}_0 \gets \{\underbrace{\beta_0^{min}, \dots, \beta_0^{min}}_{\times k}, \underbrace{\beta_0^{min}+1}_{\times k}, \dots, \underbrace{\beta_0^{max}}_{\times k}\}$
    \State initialize output list $P \gets []$
    \ForAll{$\beta_0 \in \mathfrak{B}_0$}
      \State $g_{max} \gets \min\big(G^{max},\; \beta_0 \cdot g^{max}\big)$
      \State $\mathrm{accepted} \gets \mathrm{false}$
      \While{not $\mathrm{accepted}$}
        \State sample $s \sim \mathcal{U}\llbracket 0, G^{max} \rrbracket$
        \If{$s \le g_{\max}$}
          \State $\mathrm{accepted} \gets \mathrm{true}$
        \EndIf
      \EndWhile
      \State append $(\beta_0, s)$ to $P$
    \EndFor
    \Statex
    \Return $P$
  \end{algorithmic} 
\end{algorithm}
\begin{algorithm}
  \caption{\textsc{Enumerate-Solutions}: Here $a$ and $b$ represent the number of components and the total genus respectively. Given an input configuration, previously determined (\cref{alg:donut-labels}), we enumerate all possible decompositions of the total genus into per-component genera. We further randomly pick one of them to actually create sample.\\
  \textbf{Input:} $a$, $b$, $g^{max}$ \\
  \textbf{Output:} $S$} \label{alg:donut-enum-sol}
  \begin{algorithmic}[]
      \State $S \gets \emptyset$ \Comment{Initialize solution set}
      \If{$b < 0$ \textbf{ or } $b > g^{max} \cdot a$}
          \State \Return $S$
      \EndIf
      \State \Call{Backtrack}{$a, b, 0, \emptyset, S$} \Comment{Start recursive enumeration}
      \State \Return $S$
  \end{algorithmic}
\end{algorithm}
\begin{algorithm}
  \caption{\textsc{Backtrack}: Exploring all the possible decompositions of the total genus into per-component genera boils down to a tree search problem with backtracking once we have reached the maximum genus. \\
  \textbf{Input:} $r_{\text{count}}$, $r_{\text{sum}}$, $k$, $\mathbf{x}$, $S$ \\
  \textbf{Output:} $S$} \label{alg:donut-backtrack}
  \begin{algorithmic}[]
      \If{$k > g_{\max}$} \Comment{Base case: all template types processed}
          \If{$r_{\text{count}} = 0$ \textbf{ and } $r_{\text{sum}} = 0$}
              \State $S \gets S \cup \{\mathbf{x}\}$
          \EndIf
          \State \Return
      \EndIf
      \Comment{Calculate upper bound for current template type}
      \If{$k = 0$}
          \State $u_k \gets r_{\text{count}}$
      \Else
          \State $u_k \gets \min(r_{\text{count}}, \lfloor r_{\text{sum}} / k \rfloor)$
      \EndIf
      \Comment{Try all feasible counts for template type $k$}
      \For{$n_k = 0$ \textbf{ to } $u_k$}
          \State $\mathbf{x}' \gets \mathbf{x} \cup \{n_k\}$
          \State \Call{Backtrack}{$r_{\text{count}} - n_k,\; r_{\text{sum}} - k \cdot n_k,\; k+1,\; \mathbf{x}',\; S$}
      \EndFor
  \end{algorithmic}
\end{algorithm}

\subsubsection{Shape generation}
\label{ssec:donut-shape-generation}

Each component belongs to one of three categories: superquadrics, $k$-tori, or cones. We generate each family independently.

\paragraph{Superquadrics.} We employ superellipsoids and supertoroids. Starting from a sphere or torus mesh generated with \texttt{Trimesh}, we apply the standard parametric deformation:

\begin{equation}
\text{Ellipsoid} \quad
\begin{cases}
x(u,v) &= s_x \, C_{\epsilon_1}(v) \, C_{\epsilon_2}(u) \\
y(u,v) &= s_y \, S_{\epsilon_1}(v) \, S_{\epsilon_2}(u) \\
z(u,v) &= s_z \, S_{\epsilon_1}(v)
\end{cases}
\end{equation}
\vspace{-0.2em}
\begin{equation}
    \text{Toroid} \quad
\begin{cases}
x(u,v) &= s_x \, \bigl(R + C_{\epsilon_1}(v)\bigr) \, C_{\epsilon_2}(u) \\
y(u,v) &= s_y \, \bigl(R + S_{\epsilon_1}(v)\bigr) \, S_{\epsilon_2}(u) \\
z(u,v) &= s_z \, S_{\epsilon_1}(v)
\end{cases}
\end{equation}

\noindent where $(s_x,s_y,s_z)$ are scale factors, $(\epsilon_1,\epsilon_2)$ control shape sharpness, and $C_\epsilon(\cdot), S_\epsilon(\cdot)$ denote exponentiated trigonometric functions:

\begin{equation}
\begin{aligned}
C_\epsilon(u) &= \operatorname{sign}(\cos(u)) \, |\cos(u)|^\epsilon \\
S_\epsilon(u) &= \operatorname{sign}(\sin(u)) \, |\sin(u)|^\epsilon
\end{aligned}
\end{equation}

\begin{figure}[t]
  \centering
  \includegraphics[width=1.0\linewidth]{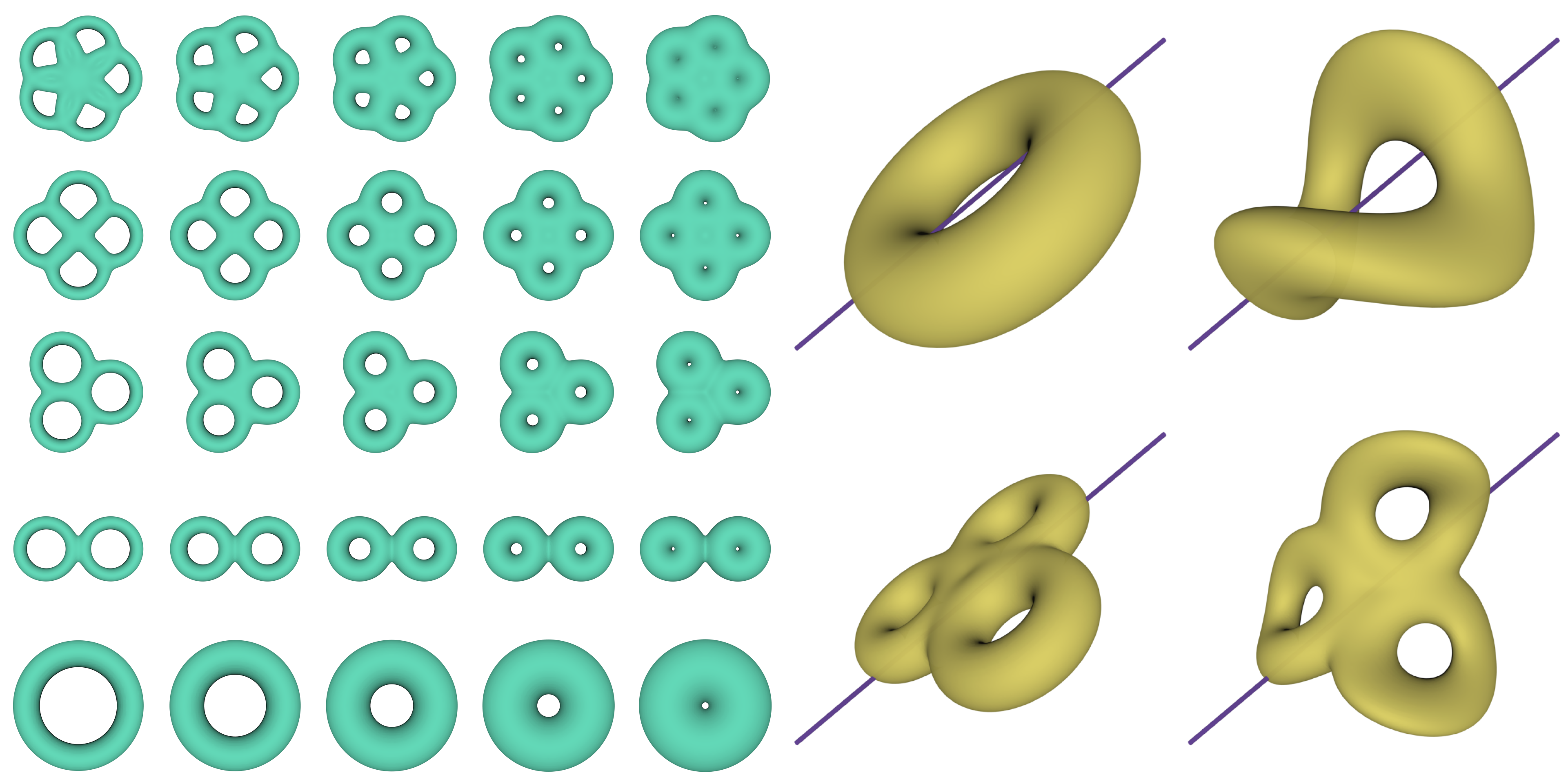}
  \caption{\textit{(left)} Examples of $k$-tori for $k\in\{1,\dots,5\}$.  
           \textit{(right)} Twisting applied to 1- and 3-tori.}
  \label{fig:donut-ktori} 
\end{figure}

\tightpara{$\mathbf{k}$-tori.} Since no closed parametric form exists for a torus with $k$ holes, we construct them via signed distance functions (SDFs). We generate $k$ individual torus SDFs, combine them using the softmin,

\begin{equation}
\operatorname{softmin}_k(s_1, s_2, \dots, s_n) 
= -\frac{1}{k} \log \left( \sum_{i=1}^n e^{-k s_i} \right)
\label{eq:softmin}
\end{equation}

and extract the final mesh using marching cubes (Fig.~\ref{fig:donut-ktori}).

\tightpara{Cones.} Cone meshes are obtained directly using \texttt{Trimesh}.

\subsubsection{Samples variety}

To avoid geometric bias, we randomize all shape hyperparameters (e.g., scales, superquadric exponents, major/minor radii) within predefined ranges. We further apply random rigid motions and twisting deformations (Fig.~\ref{fig:donut-ktori}, right) to each component before merging.

\subsection{Baselines}
\label{ssec:suppl-baselines}

\paragraph{Implementation.} All baselines are trained from scratch on DONUT (Table~\ref{tab:donut-baselines}) using their official implementations and hyperparameters.

\tightpara{Training.} We train every model for 200 epochs with batch size 32 using Adam with initial learning rate $10^{-3}$, reduced by a factor~0.5 every 20 epochs. We apply random rotations, translations, and scaling.

\subsection{Implementation and training of FILTR}
\label{ssec:filtr-implem}

\subsubsection{Input processing} 

All experiments use point clouds subsampled to 1024 points and normalized within the unit sphere. Persistence diagrams and all topological computations are performed using \texttt{Gudhi} \cite{gudhi:urm}.

\subsubsection{Pretrained encoders} 
All encoders follow the same architecture and differ only by their pretraining method \cite{yu2022pointbert,pang2022pointmae,chen2023pointgpt,zhang2024pcp}.  
Point clouds are partitioned into 64 patches of 32 points. Each patch is embedded into a 384-dimensional vector via a shared MLP, and patch centroids are mapped to positional encodings through another MLP. A 12-block transformer processes the resulting sequence. We use the pretrained checkpoints released by the respective authors.

\subsubsection{Adapter} 
We map encoder outputs to the decoder space by applying layer normalization followed by a linear projection from 384 to 256 dimensions. Positional encodings are projected separately with a linear layer.

\subsubsection{Transformer decoder} 
We adopt the DETR decoder architecture: 6 transformer blocks with self- and cross-attention, hidden dimension 256, and 8 attention heads.

\subsubsection{Baselines for FILTR}
For PointNet and DGCNN, we extract the per-point features prior to their global pooling stage and feed them to a 3-block transformer encoder with hidden dimension 256. For PointNet++ and RepSurf, whose architectures progressively downsample the point cloud through pooling, we instead use the features obtained after the second PointNet Set Abstraction layer (128 points). Using only the final globally pooled feature vector from the third abstraction layer led to unstable training. Positional embeddings are computed using an MLP: from each point of the input point cloud for PointNet and DGCNN, and from the pooled 128-point representation for PointNet++ and RepSurf. Table~\ref{tab:filtr-trainable} reports the parameter counts for all FILTR variants.

\subsubsection{Training}
Models are trained on 23\,579 DONUT point clouds using a single NVIDIA L40 GPU. We use batch size~64, train for 250 epochs with 5 warm-up epochs, and apply cosine learning-rate decay.  
We use $N=250$ queries and optimize using AdamW with initial learning rate $10^{-4}$. Loss weights are $\mu_{\text{recon}}=1.0$, $\mu_{\text{exist}}=0.1$, $\mu_{\text{diag}}=0.1$; matching costs use $\lambda_{\text{reg}}=1.0$ and $\lambda_{\text{exist}}=0.1$. No data augmentation is applied. Pretrained encoders remain frozen; baseline models are trained end-to-end (Fig.~\ref{fig:filtr-baseline}).

\begin{table}[h]
\centering
\begin{tabular}{lc}
\toprule
Model & \#params $(\times 10^6)$ \\
\midrule
FILTR + pretrained   & 5.7 \\
FILTR + PointNet & 8.1 \\
FILTR + DGCNN    & 8.7 \\
FILTR + PointNet++  & 7.9 \\
FILTR + RepSurf  & 7.9 \\
\bottomrule
\end{tabular}
\caption{\textbf{Number of trainable parameters for FILTR with different encoders.} Training an end-to-end pipeline adds around 2.5 million parameters compared to using a frozen pretrained encoder.}
\label{tab:filtr-trainable}
\end{table} 
\section{3D vs. latent prediction pretraining}
\label{sec:recon-demo}

In this work, we focus on encoders pretrained with a 3D reconstruction objective. This choice is motivated by the geometric guarantees naturally provided by optimizing spatial reconstruction metrics.

\subsection{Theoretical justification}
\label{ssec:th-demo}

Let $X$ and $\hat{X} \in \mathbb{R}^{N\times 3}$ be the ground-truth and reconstructed point clouds. 3D-prediction encoders minimize the mean Chamfer distance ($CD$) between $X$ and $\hat{X}$. To connect this objective to topological stability, we first relate $CD$ to the Hausdorff distance ($d_H$), which measures the maximum spatial discrepancy between the two point sets. 

Because the unreduced sum of minimum distances upper-bounds the maximum error, the Hausdorff distance is bounded by the mean Chamfer distance scaled by the number of points $N$:
\begin{equation}
    d_H(X, \hat{X}) \leq N \cdot CD(X, \hat{X})
\end{equation}

Furthermore, the stability theorem for persistence diagrams establishes that the Bottleneck distance ($d_B$) between the persistence diagrams $D(X)$ and $D(\hat{X})$ is bounded by the Hausdorff distance:
\begin{equation}
    d_B (D(X), D(\hat{X})) \leq d_H(X, \hat{X})
\end{equation}

Combining these inequalities yields:
\begin{equation}
    d_B (D(X), D(\hat{X})) \leq N \cdot CD(X, \hat{X})      
\end{equation}

Therefore, minimizing the Chamfer distance explicitly bounds the topological error. This mathematical guarantee ensures that learned features optimized for 3D reconstruction carry sufficient information to reconstruct persistence diagrams. In contrast, latent-prediction methods lack this geometric constraint. \textit{Note:} The bounds derived above hold for the $\alpha$-filtration, but the general results still hold for the Vietoris-Rips filtration, albeit with a different constant factor.

\subsection{Results on Point2Vec}
\label{ssec:res-p2vec}

\begin{figure}[h]
  \centering
  \includegraphics[width=1.0\linewidth]{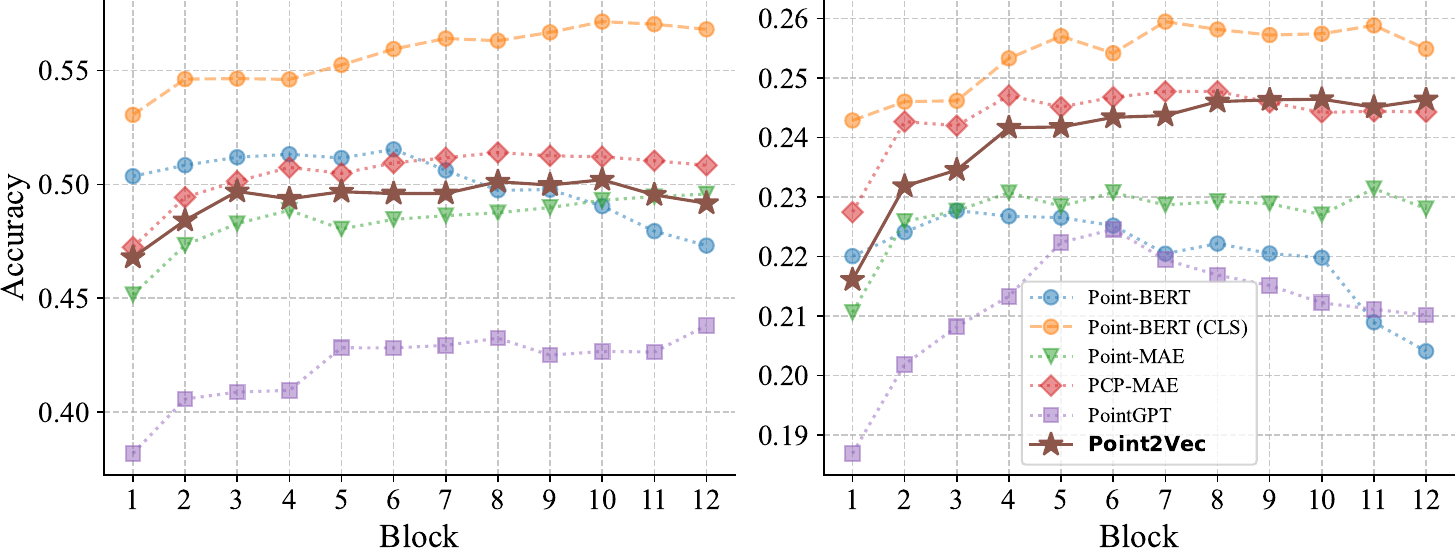}
  \caption{\textbf{Probing results on Point2Vec, compared to other encoders.} Point2Vec follows the same trend as other encoders.}
   \label{fig:probing-p2vec} 
\end{figure}

\begin{figure}[h]
  \centering
  \includegraphics[width=1.0\linewidth]{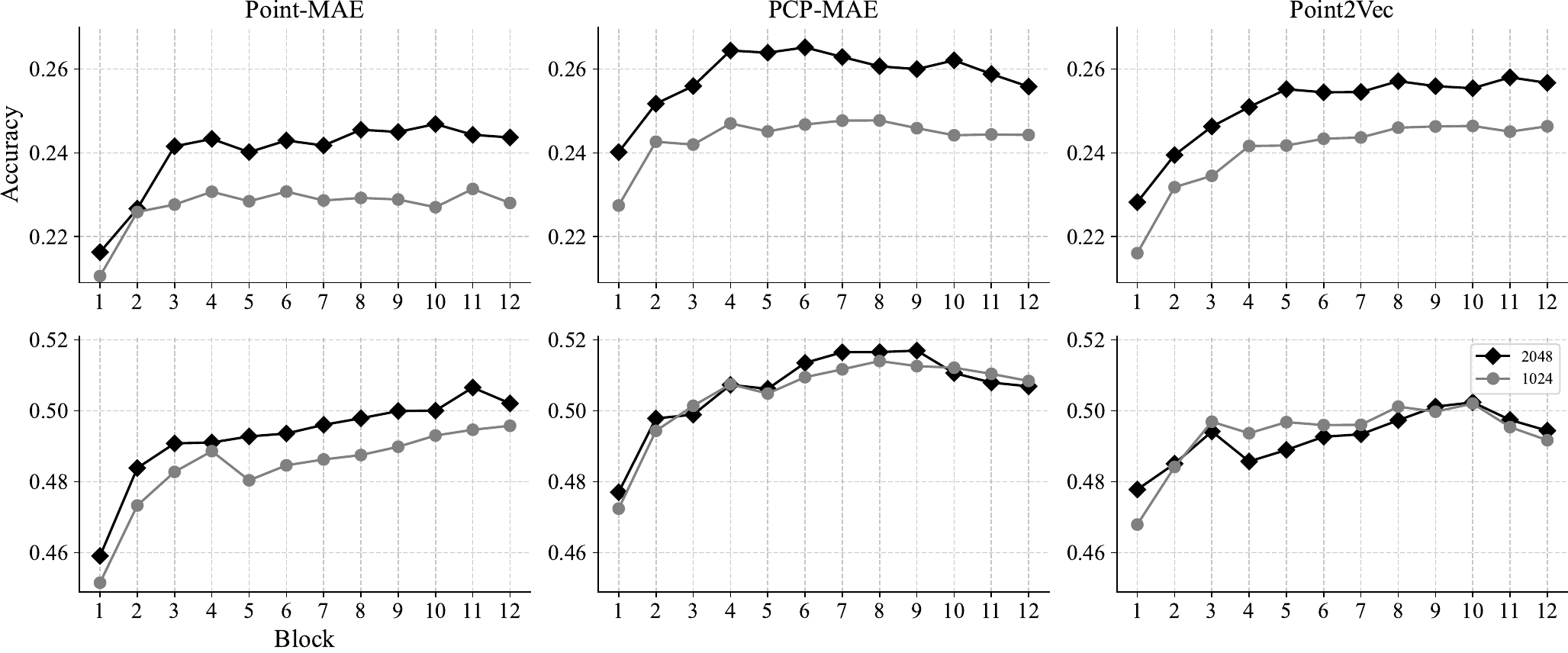}
  \caption{\textbf{Probing with different point cloud densities.} We report probing accuracies for Point-MAE, PCP-MAE, and Point2Vec on features computed from 1024- and 2048-point clouds. (top row) genus, (bottom row) connected components.}
   \label{fig:probing-density} 
\end{figure}

\begin{figure}[h]
  \centering
  \includegraphics[width=1.0\linewidth]{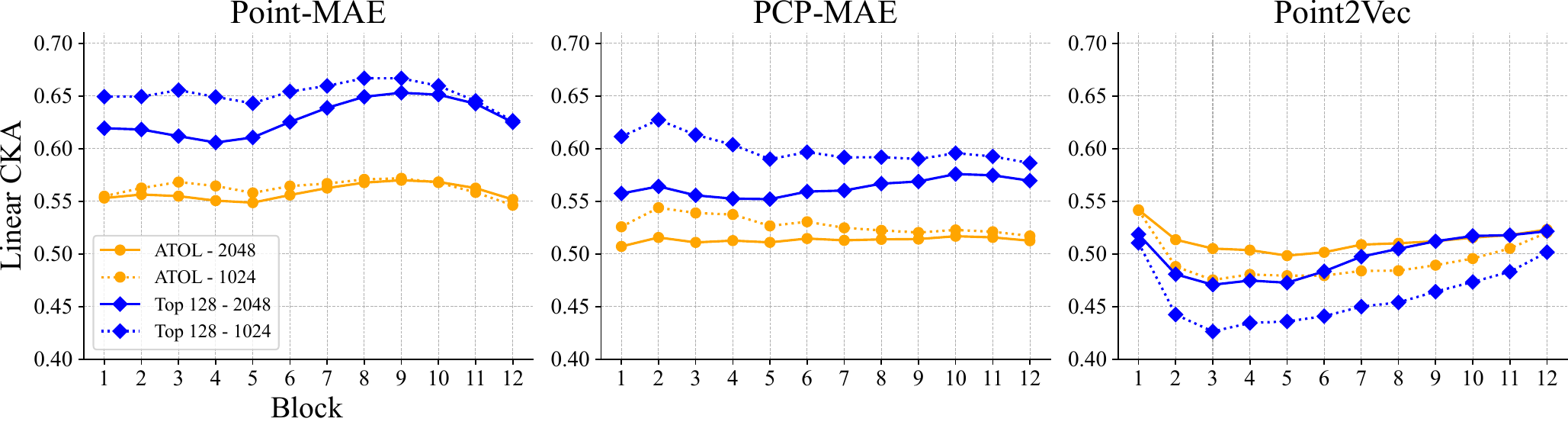}
  \caption{\textbf{CKA results with different point cloud densities.} We report alignment scores for Point-MAE, PCP-MAE, and Point2Vec on features computed from 1024- and 2048-point clouds.}
   \label{fig:cka-density} 
\end{figure}

Figure \ref{fig:probing-p2vec} highlights that probing results on Point2Vec features are comparable with Point-MAE and PCP-MAE. This indicates that encoders, regardless of their pretraining objectives (3D or latent prediction), capture a similar amount of global structural information. However, alignment scores in Figure \ref{fig:cka-density} show that Point2Vec lags behind its 3D reconstruction-based counterparts. This suggests that latent-prediction encoders struggle to capture local topology, which strictly aligns with the theoretical guarantees discussed in \cref{ssec:th-demo}.
\section{Experiments}

\subsection{Per-category probing results}
\label{ssec:donut-probing-results}

\begin{table*}[t]
\centering
\resizebox{\textwidth}{!}{
\begin{tabular}{l|ccccccccccc|cccccc}
\toprule
Model & \multicolumn{11}{c|}{\textbf{Genus}} & \multicolumn{6}{c}{\textbf{Connected Components}} \\
& 0 & 1 & 2 & 3 & 4 & 5 & 6 & 7 & 8 & 9 & 10 & 1 & 2 & 3 & 4 & 5 & 6 \\
% \cmidrule{lr}{2-4} \cmidrule{lr}{5-7}
\midrule
\multicolumn{1}{l}{} & \multicolumn{17}{c}{\textit{Pretrained-Frozen encoders}} \\
\midrule
Point-BERT \cite{yu2022pointbert} & \cellyel $53.3_{(9)}$ & $27.2_{(7)}$ & \cellyel $27.5_{(3)}$ & \cellyel $19.0_{(3)}$ & \cellyel $17.7_{(10)}$ & \cellyel $23.3_{(3)}$ & \cellyel $12.8_{(5)}$ & \cellyel $12.0_{(7)}$ & \cellorg $16.2_{(5)}$ & $20.7_{(1)}$ & $20.8_{(5)}$ & \cellyel $84.1_{(3)}$ & \cellorg $61.3_{(3)}$ & \cellorg $43.7_{(4)}$ & \cellred $\mathbf{34.8_{(6)}}$ & \cellorg $31.8_{(10)}$ & \cellred $\mathbf{55.8_{(6)}}$ \\
Point-MAE \cite{pang2022pointmae} & \cellorg$56.2_{(10)}$ & \cellred $\mathbf{33.9_{(5)}}$ & \cellorg $28.3_{(7)}$ & \cellorg $19.3_{(6)}$ & \cellred $\mathbf{19.8_{(3)}}$ & \cellorg $23.5_{(2)}$ & \cellred $\mathbf{13.0_{(8)}}$ & \cellorg $13.2_{(12)}$ & $15.8_{(1)}$ & \cellyel $22.0_{(10)}$ & \cellyel $21.3_{(7)}$ & \cellorg $84.2_{(4)}$ & \cellyel $60.0_{(12)}$ & \cellyel $40.2_{(5)}$ & \cellyel $33.0_{(8)}$ & \cellyel $30.7_{(11)}$ & \cellorg $53.9_{(9)}$ \\
PointGPT \cite{chen2023pointgpt}  & $51.6_{(12)}$ & \cellyel $29.0_{(12)}$ & $27.0_{(10)}$ & $17.8_{(6)}$ & $17.7_{(10)}$ & $22.2_{(12)}$ & $10.7_{(2)}$ & $11.9_{(5)}$ & \cellyel $16.0_{(10)}$ & \cellorg $22.1_{(6)}$ & \cellorg $22.5_{(3)}$ & $77.6_{(12)}$ & $50.6_{(12)}$ & $34.1_{(12)}$ & $25.0_{(8)}$ & $30.1_{(6)}$ & $48.9_{(12)}$ \\
PCP-MAE \cite{zhang2024pcp}      & \cellred $\mathbf{56.6_{(4)}}$ & \cellorg$33.4_{(5)}$ & \cellred $\mathbf{30.7_{(10)}}$ & \cellred $\mathbf{20.3_{(3)}}$ & \cellorg $19.6_{(7)}$ & \cellred $\mathbf{26.0_{(10)}}$ & \cellorg $12.9_{(2)}$ & \cellred $\mathbf{15.8_{(3)}}$ & \cellred $\mathbf{18.0_{(5)}}$ & \cellred $\mathbf{23.7_{(4)}}$ & \cellred $\mathbf{23.3_{(5)}}$ & \cellred $\mathbf{86.0_{(8)}}$ & \cellred $\mathbf{64.3_{(7)}}$ & \cellred $\mathbf{44.1_{(8)}}$ & \cellorg $34.4_{(12)}$ & \cellred $\mathbf{33.6_{(5)}}$ & \cellyel $53.6_{(7)}$ \\
\midrule
\multicolumn{1}{l}{} & \multicolumn{17}{c}{\textit{Baseline models trained from scratch}} \\
\midrule
PointNet \cite{Qi2017PointNet}   & 55.7 & 30.8 & 13.4 & 8.00 & 8.92 & 23.1 & 4.18 & 1.40 & 5.88 & \cellred \bf 46.9 & 13.3 & 89.1 & 71.5 & 41.4 & 31.7 & 26.9 & 52.1 \\
PointNet++ \cite{Qi2017PointNetPlusPlus} & \cellorg 89.8 & \cellorg 63.5 & \cellorg 64.7 & \cellorg 55.6& \cellorg 52.0 & \cellorg 50.4 & \cellorg 22.2 & \cellorg 32.1 & \cellorg 25.3 & \cellorg 38.0 & \cellorg 53.8 & \cellorg 99.6 & \cellorg 94.5 & \cellyel 81.7 & \cellyel 61.4 & \cellyel 47.5 & \cellyel 65.9 \\
DGCNN \cite{Wang2019DGCNN}      & \cellyel 80.0 & \cellorg 63.5 & \cellyel 47.7 & \cellyel 46.3 & \cellyel 33.2 & \cellyel 37.4 & \cellyel 13.8 & \cellyel 21.4 & \cellyel 10.4 & 28.1 & \cellyel 52.4 & \cellorg 99.6 & \cellyel 93.9 & \cellorg 83.3 & \cellorg 71.3 & \cellred \bf 59.2 & \cellorg 72.7 \\
RepSurf \cite{Ran2022RepSurf}      & \cellred \bf 93.0 & \cellred \bf 70.3 & \cellred \bf 77.5 & \cellred \bf 64.5 & \cellred \bf 64.0 & \cellred \bf 54.7 & \cellred \bf 30.5 & \cellred \bf 29.8 &  \cellred \bf 26.7 & \cellyel 32.6 & \cellred \bf 63.1 & \cellred \bf  100 & \cellred \bf 97.4 &  \cellred \bf 89.3 & \cellred \bf 74.4 & \cellorg 54.8 & \cellred \bf 82.8 \\
\bottomrule
\end{tabular}
}
\caption{\textbf{Performance per category on DONUT.} We report classification accuracies (\%) for genus and number of connected components prediction, for pretrained encoders and baseline models trained from scratch on DONUT. For pretrained encoders, we indicate in subscript the transformer block that achieved the best accuracy.}
\label{tab:donut-probing}
\end{table*}

Table~\ref{tab:donut-probing} reports per-category probing accuracies along with baseline results. As expected, accuracy generally decreases for categories with higher topological complexity. Although Fig.~\ref{fig:probing-results} shows that probing performance tends to improve in deeper transformer blocks, the depth of the best-performing block (indicated in subscript) does not exhibit a consistent relationship with category difficulty. Finally, RepSurf~\cite{Ran2022RepSurf} clearly outperforms all models trained from scratch, suggesting that explicitly encoding surface-based features provides a substantial advantage for capturing the underlying topology of point clouds.

\subsection{Additional results on denser point-clouds}
\label{ssec:donut-probing-results-denser}

We discuss how probing and CKA results change when using encoder features computed from 2048-point clouds (instead of 1024). Intuitively, denser point clouds carry more information about the global structure of the shape, by "filling" space between points of 1024-point clouds. Therefore, fine topological structures are more salient. Figure \ref{fig:probing-density} shows that genus prediction accuracy benefits from denser point-clouds, while connected components prediction remains similar. It is expected, since detecting connected components relies less on how densely sampled the shape is. This also reveals that, despite yielding rather poor probing scores, encoders implicitly carry some topological information about higher-order structures, that can be disambiguated with denser point clouds.

\subsection{Relevance of CKA scores}
\label{ssec:cka-relevance}

\begin{figure}
  \centering
  \includegraphics[width=1.0\linewidth]{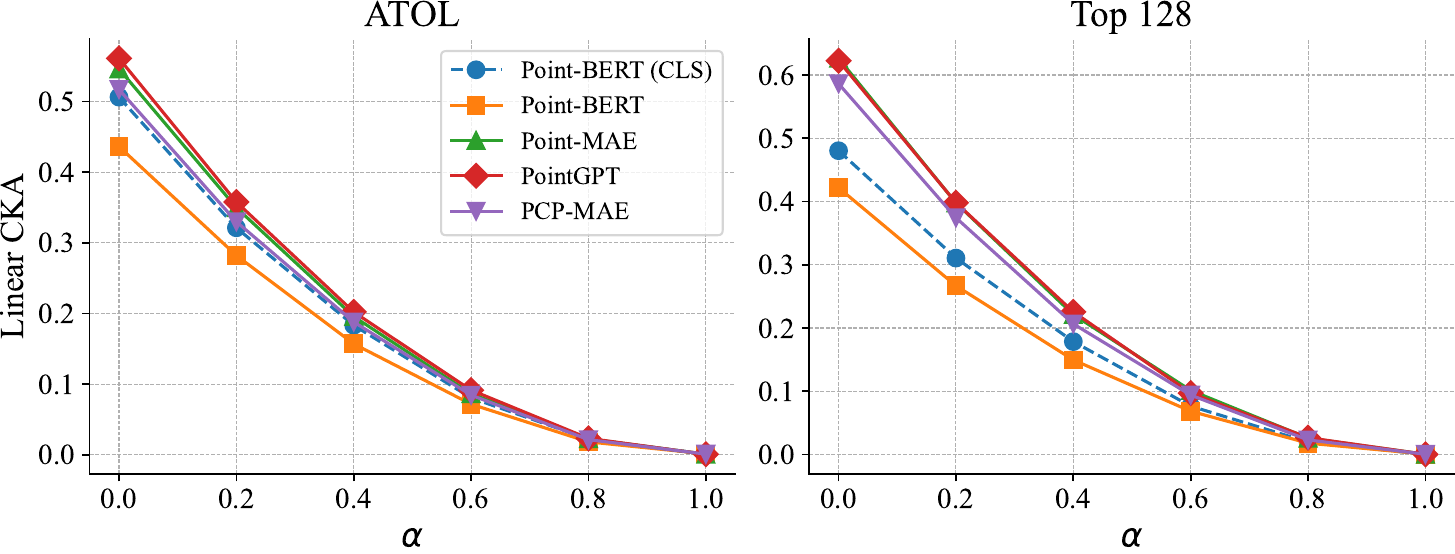}
  \caption{\textbf{CKA under controlled feature mismatch.} CKA similarity between the last transformer block of each encoder and ATOL/top-128 vectorizations on DONUT. A fraction $\alpha$ of features is randomly permuted, and results are averaged over 3 runs.}
   \label{fig:cka-ablation} 
\end{figure}

The CKA similarities in Figure \ref{fig:cka-results} allow comparison between encoders, but do not directly indicate whether the absolute CKA values represent meaningful alignment. Because CKA can be influenced by feature dimensionality and background correlations, we validate its interpretability through a controlled perturbation.

Let $\{f_i\}_{i=1}^n$ denote the features extracted by the encoder and $\{v_i\}_{i=1}^n$ the corresponding vectorized persistence descriptors, with a one-to-one correspondence between indices. For a given proportion $\alpha\in[0,1]$, we introduce a permutation $\sigma^{(\alpha)}$ that randomly permutes a fraction $\alpha$ of the indices and therefore creates mismatches. We then compute $\text{CKA}(f_{\sigma^{(\alpha)}(i)}, v_i)$ as a function of $\alpha$.

Figure~\ref{fig:cka-ablation} shows the resulting degradation for ATOL and top-128 vectorizations. The rapid decline in similarity confirms that high CKA values cannot be explained by dimensionality alone and instead reflect genuine structural alignment between learned features and persistence information.

\subsection{Additional results on FILTR}
\label{ssec:add-filtr}

\begin{figure}
  \centering
  \includegraphics[width=1.0\linewidth]{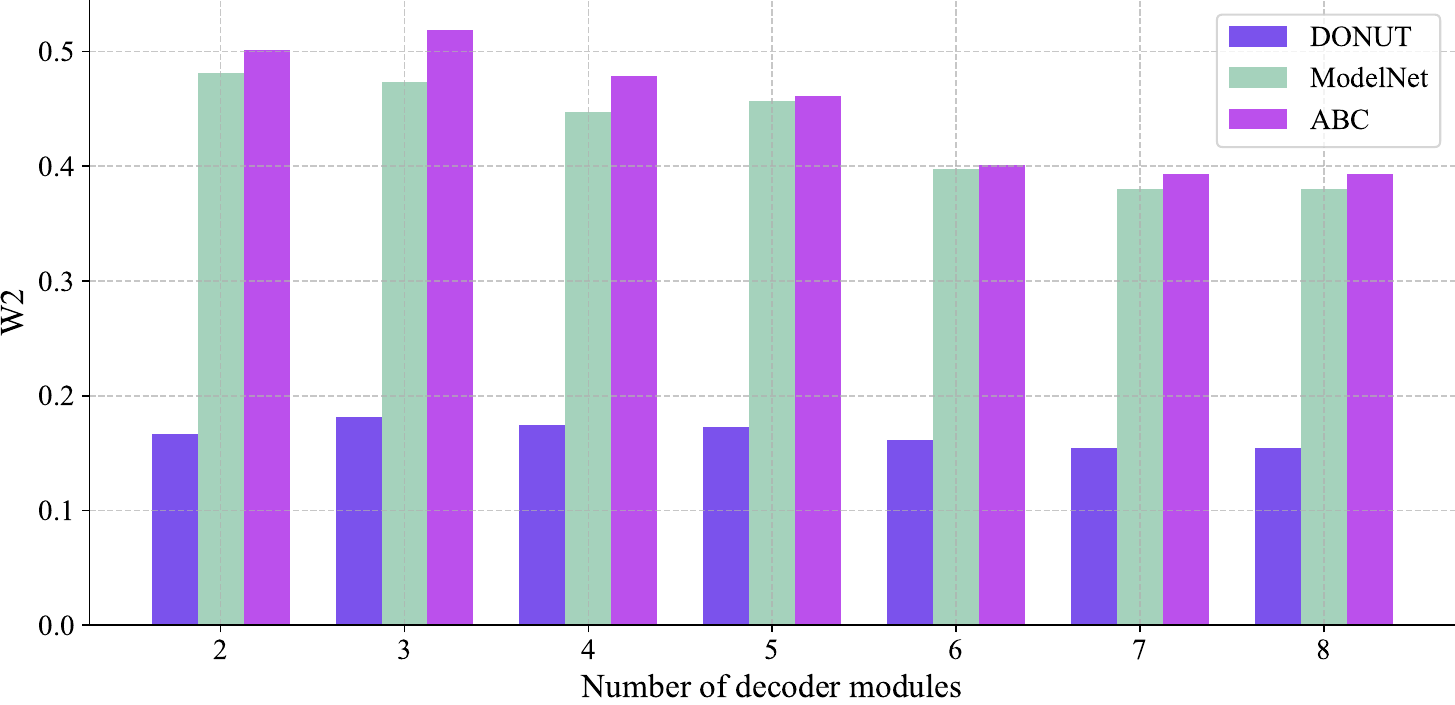}
  \caption{\textbf{Effect of decoder depth.} We train FILTR on DONUT with varying decoder depth using a Point-MAE backbone. We report 2-Wasserstein distances on DONUT (test), ModelNet, and ABC.}
   \label{fig:filtr_dec} 
\end{figure}

\begin{table}[h]
\centering
\resizebox{\linewidth}{!}{
\begin{tabular}{l|cccc}
\toprule
% \cmidrule{lr}{2-4} \cmidrule{lr}{5-7}
Model & FLOPS ($\times 10^9$) & Allocated Memory (GB) & Training (hours) & Inference (ms) \\
\midrule
FILTR + pretrained   & 4.09 & 2.36 & 9 & 25.5$\pm$ 0.1 \\
FILTR + PointNet   & 14.64 & 11.43 & 19 & 140$\pm$ 0.1 \\
FILTR + DGCNN   & 19.10 & 15.45 & 22 & 230$\pm$ 0.2 \\
\bottomrule
\end{tabular}
}
\caption{\textbf{Computational Cost.} FLOPS are estimated on a single input sample. Training setup is similar to the one used in the main paper. Inference time is estimated for a batch size of 64.}
\label{tab:comp-cost}
\vspace{-0.35cm}
\end{table}

\paragraph{Decoder depth.} To evaluate the role of decoder depth, we train FILTR with different numbers of transformer decoder blocks while keeping all other hyperparameters fixed. As shown in Fig.~\ref{fig:filtr_dec}, performance on ModelNet and ABC improves up to six decoder blocks, after which additional depth yields diminishing returns.

\begin{table}[h]
\centering
\resizebox{\linewidth}{!}{
\begin{tabular}{l|ccc}
\toprule
% \cmidrule{lr}{2-4} \cmidrule{lr}{5-7}
Loss & $W_{2\,(\times 10^{-2})}$ & $d_{B\,(\times 10^{-3})}$ & PIE \\
\midrule
$\mathcal{L}_{\text{recon}}$   & 23.63 & 10.47 & 4.079 \\
\midrule
$\mathcal{L}_{\text{recon}} + \mathcal{L}_{\text{exist}}$ w/ $p_e$ & 16.42 & 9.866 & 1.193 \\
$\mathcal{L}_{\text{recon}} + \mathcal{L}_{\text{exist}}$ w/o $p_e$ & 43.06 & 10.75 & \cellgra -- \\
\midrule
$\mathcal{L}_{\text{recon}} + \mathcal{L}_{\text{exist}} + \mathcal{L}_{\text{diag}}$ w/ $p_e$ & 17.27 & 9.917 & 1.107 \\
$\mathcal{L}_{\text{recon}} + \mathcal{L}_{\text{exist}} + \mathcal{L}_{\text{diag}}$ w/o $p_e$ & 17.24 & 9.918 & \cellgra -- \\
\bottomrule
\end{tabular}
}
\caption{\textbf{Ablation study of losses.} We use a Point-MAE encoder that achieves competitive results on DONUT (Tab. \ref{tab:filtr-recon}). Similar results are observed with other encoders (see \cref{tab:add-filtr-ablation-loss}). We report results for both thresholded (w/ $p_e$) and non-thresholded (w/o $p_e$) existence probability $p_e$ when using the existence loss $\mathcal{L}_{\text{exist}}$.}
\label{tab:filtr-ablation-loss}
\vspace{-0.2cm}
\end{table}
\begin{table}[h]
\centering
\resizebox{\linewidth}{!}{
\begin{tabular}{l|ccc}
\toprule
% \cmidrule{lr}{2-4} \cmidrule{lr}{5-7}
Loss & Point-BERT & PCP-MAE & PointGPT \\
\midrule
$\mathcal{L}_{\text{recon}}$   & 31.67 & 26.62 & 53.73 \\
\midrule
$\mathcal{L}_{\text{recon}} + \mathcal{L}_{\text{exist}}$ w/o $p_e$ & 113.6 & 67.64 & 291.72 \\
$\mathcal{L}_{\text{recon}} + \mathcal{L}_{\text{exist}}$ w/ $p_e$ & \cellyel 17.33 & \cellred \bf 17.02 & \cellyel 18.20 \\
\midrule
$\mathcal{L}_{\text{recon}} + \mathcal{L}_{\text{exist}} + \mathcal{L}_{\text{diag}}$ w/o $p_e$ & \cellorg 16.23 & \cellyel 17.63 & \cellorg 17.95 \\
$\mathcal{L}_{\text{recon}} + \mathcal{L}_{\text{exist}} + \mathcal{L}_{\text{diag}}$ w/ $p_e$ & \cellred \bf 16.18 & \cellorg 17.18 & \cellred \bf 17.86 \\
\bottomrule
\end{tabular}
}
\caption{\textbf{Ablation study of losses on additional encoders.} We report $W_{2\,(\times 10^{-2})}$ for FILTR trained with different pretrained encoders. We report results for both thresholded (w/ $p_e$) and non-thresholded (w/o $p_e$) existence probability $p_e$ when using the existence loss $\mathcal{L}_{\text{exist}}$.}
\label{tab:add-filtr-ablation-loss}
\vspace{-0.2cm}
\end{table}

\tightpara{Ablations.} Table \ref{tab:filtr-ablation-loss} shows that adding $\mathcal{L}_\textbf{exist}$ substantially improves reconstruction, compared to $\mathcal{L}_\textbf{recon}$ alone. As expected, without $\mathcal{L}_\textbf{diag}$, it is essential to threshold non-existing persistence pairs to get good performance. Finally, while introducing $\mathcal{L}_\textbf{diag}$ slightly reduces performance for $W_2$, they remain identical with and without thresholding. We also note that overall, the Bottleneck distance $d_B$ seems to be hardly impacted by these variants. Furthermore, Table~\ref{tab:add-filtr-ablation-loss} extends the loss ablation experiments by reporting 2-Wasserstein distances for all remaining encoders.

\begin{table}
\centering
\begin{tabular}{l|ccc}
\toprule
% \cmidrule{lr}{2-4} \cmidrule{lr}{5-7}
Dataset &  $W_{2\,(\times 10^{-2})}$ & $d_{B\,(\times 10^{-3})} $ & PIE \\
\midrule
DONUT   & 36.17 & 10.26 & 13.26 \\
ModelNet & 56.37 & 13.07 & 14.28 \\
ABC & 73.77 & 32.15 & 5.92 \\
\bottomrule
\end{tabular}
\caption{\textbf{Reconstruction results with RepSurf.}}
\label{tab:repsurf-res}
\vspace{-0.2cm}
\end{table}

\subsubsection{Discussion on PointNet++} 
To adapt PointNet++ and RepSurf to our setting, we use the 128 per-region features produced after the second Set Abstraction layer, before the final pooling stage, as input to the FILTR decoder. These intermediate features preserve local geometric information while being stable enough to train effectively, in contrast to using the final globally pooled representation, which led to unstable training. Table \ref{tab:repsurf-res} shows the performance of FILTR with RepSurf as feature extractor.

\subsubsection{Performance of DGCNN baseline}
\begin{table}[h]
\centering
\resizebox{\linewidth}{!}{
\begin{tabular}{l|ccc}
\toprule
% \cmidrule{lr}{2-4} \cmidrule{lr}{5-7}
Extractor & $W_{2\,(\times 10^{-2})}$ & $d_{B\,(\times 10^{-3})} $ & PIE \\
\midrule
Point-MAE$_C$  & $29.74$ & $10.54$ & $3.874$ \\
DGCNN (E2E)  & $113.1_{(+83.36)}$ & $36.67_{(+26.13)}$ & $8.301_{(+4.427)}$ \\
\bottomrule
\end{tabular}
}
\caption{\textbf{Results on DONUT under low data regime (2K shapes).} The two first rows compare frozen and E2E setups.}
\label{tab:low-data}
\vspace{-0.5cm}
\end{table} 

As pointed in Section \ref{ssec:filtr_results}, the end-to-end (E2E) baseline with DGCNN feature extractor tends to outperform pretrained feature extractors (\cref{tab:filtr-recon}) on ModelNet and ABC. We hypothesize that this stems from two reasons: (1) E2E models naturally excel in \textit{high-data regimes} by overfitting to task-specific distributions. However, FILTR with frozen encoders demonstrate widely superior performance in \textit{low-data regimes}. As shown in Table \ref{tab:low-data}, FILTR significantly outperforms E2E DGCNN. (2) DGCNN's architecture is explicitly tailored to capture local topology of point clouds--as claimed by the authors--making it biased towards this task.

\subsubsection{Results with Vietoris-Rips Filtration}

While our primary experiments utilize the $\alpha$-filtration, FILTR is fundamentally agnostic to the specific choice of filtration. Because the architecture treats persistence diagrams strictly as unordered sets, it only requires the resulting persistence pairs as target inputs during training. Consequently, the model is fully capable of learning and fitting the specific data distribution of the target diagrams regardless of the underlying mathematical method used to compute them. To empirically demonstrate this flexibility, we train and evaluate (\cref{tab:vr-low-data}) FILTR on a subset of 2K samples from the DONUT dataset on Vietoris-Rips filtration.

\begin{table}[h]
\centering
\resizebox{\linewidth}{!}{
\begin{tabular}{l|ccc}
\toprule
% \cmidrule{lr}{2-4} \cmidrule{lr}{5-7}
Extractor & $W_{2\,(\times 10^{-2})}$ & $d_{B\,(\times 10^{-3})} $ & PIE \\
\midrule
Point-MAE$_C$ (Rips)  & $92.55$ & $41.23$ & $8.562$ \\
\bottomrule
\end{tabular}
}
\caption{\textbf{Results on DONUT under low data regime (2K shapes) for Vietoris-Rips filtration.} FILTR is trained on Point-MAE (combined features) to predict VR persistence diagrams. While \textit{no quantile thresholding} is applied to the predicted diagrams, the model still achieves competitive performance with the E2E baseline (\cref{tab:low-data}).}
\label{tab:vr-low-data}
\vspace{-0.5cm}
\end{table} 

\subsubsection{Qualitative results}

\begin{figure*}
  \centering
  \includegraphics[width=1.0\linewidth]{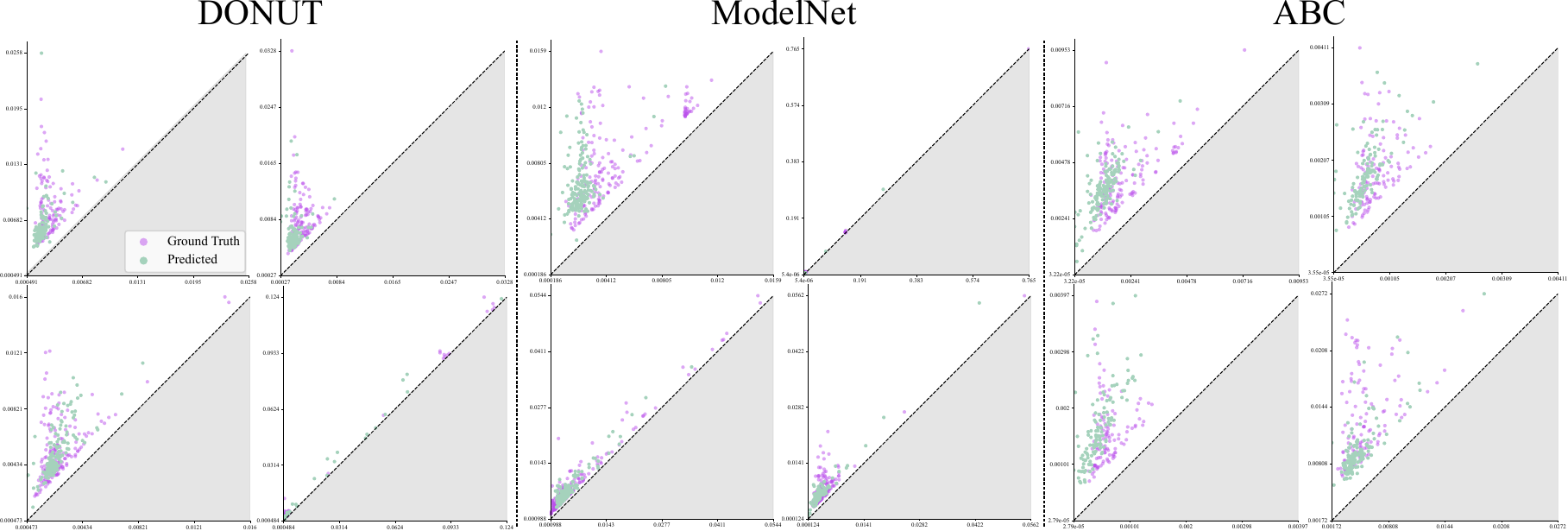}
  \caption{\textbf{Predicted persistence diagrams.} Predicted vs. ground-truth persistence diagrams from FILTR (Point-MAE backbone) on DONUT, ModelNet, and ABC samples.}
   \label{fig:res_pd} 
\end{figure*}

\begin{figure}
  \centering
  \includegraphics[width=1.0\linewidth]{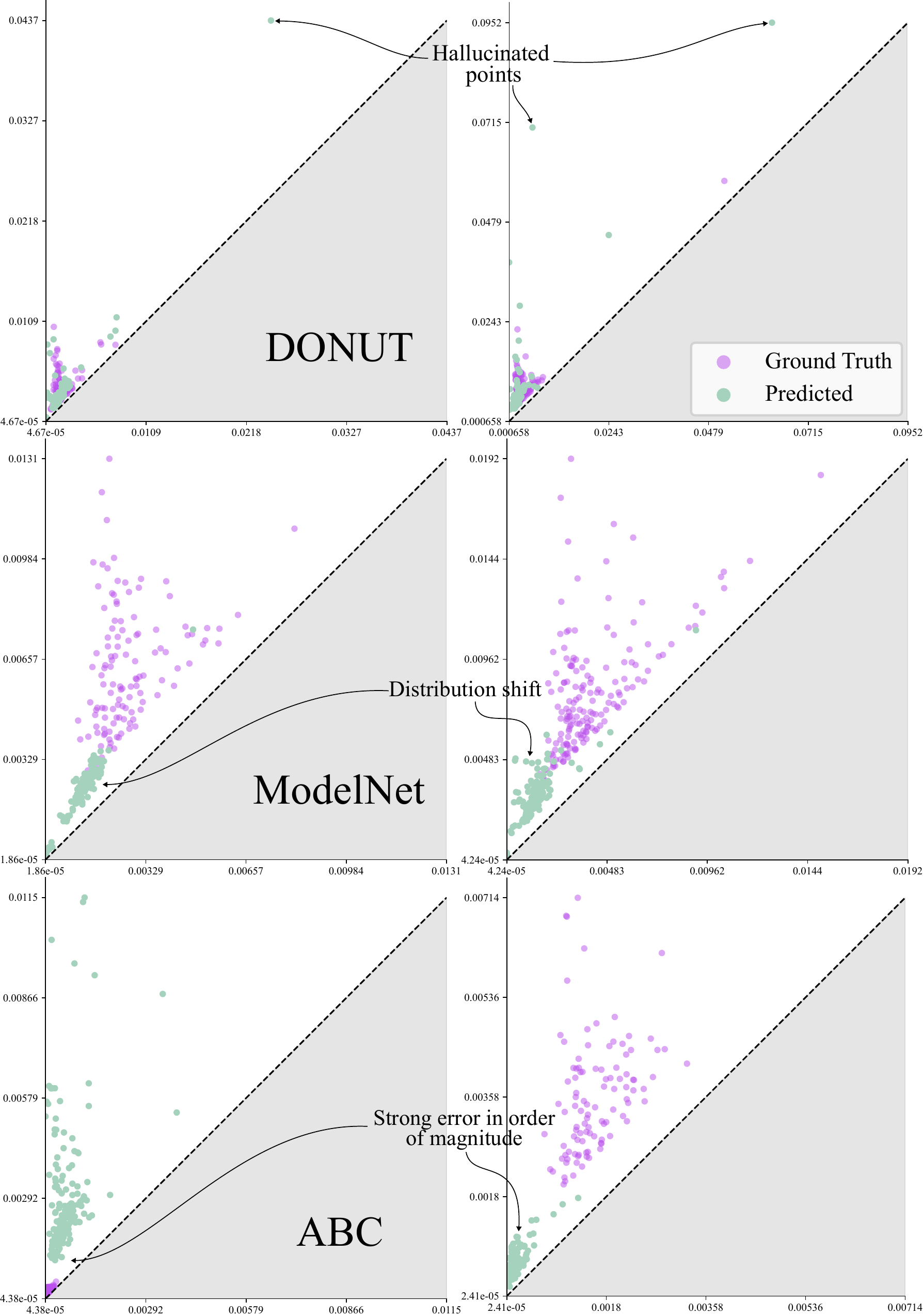}
  \caption{\textbf{Failure cases.} Predicted vs. ground-truth persistence diagrams from FILTR (Point-MAE backbone) on DONUT, ModelNet, and ABC samples.}
   \label{fig:res_pd_fail}
\end{figure}

\begin{figure}
\centering
\includegraphics[width=1.0\linewidth]{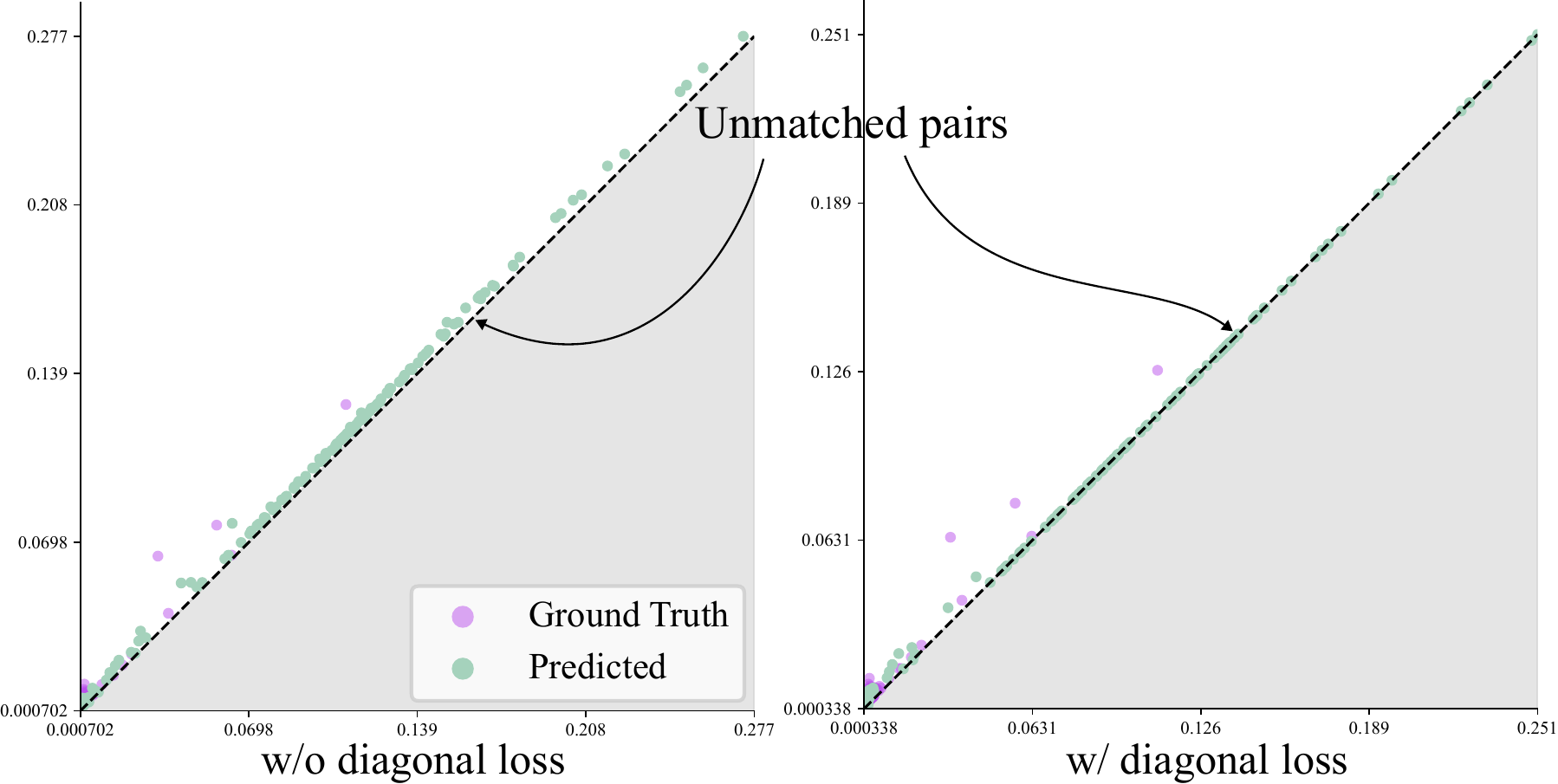}
\caption{\textbf{Effect of $\mathcal{L}_{\text{diag}}$.} \textit{(left)} Unmatched pairs a close to the diagonal but still contributing to the 2-Wasserstein distance. \textit{(right)} With the diagonal loss, unmatched pairs are exactly on the diagonal, contributing zero to the distance.}
   \label{fig:res_pd_diag}
\end{figure}

\paragraph{Reconstruction.} Figure~\ref{fig:res_pd} shows that FILTR captures the overall structure of persistence diagrams across datasets. The predicted distributions and magnitudes of persistence pairs generally align with the ground truth. However, as discussed in Section~\ref{ssec:probing}, the most persistent pairs, which correspond to the dominant topological features of a shape, remain difficult to predict accurately. Estimating these pairs requires the encoder to capture global geometric structure, a capability that pretrained models struggle with, as indicated in Table~\ref{tab:donut-baselines}.

Figure~\ref{fig:res_pd_fail} illustrates typical failure cases. The most common error is a shift between the predicted and ground-truth locations of persistence pairs. This effect appears on both ModelNet and ABC, but is more pronounced on ABC, where mismatches may span several orders of magnitude. This behavior is consistent with the distribution shift between datasets: pretrained encoders are primarily exposed to ShapeNet-like geometry during pretraining, while ABC shapes exhibit topological configurations that are not well represented in ShapeNet.

\tightpara{Effect of the diagonal loss.} Figure~\ref{fig:res_pd_diag} illustrates the impact of including the diagonal loss term $\mathcal{L}_{\text{diag}}$ in FILTR's training objective. Without this term, the model tends to produce persistence diagrams with a higher density of low-persistence points near the diagonal. They require using the existence probability to filter noisy points and retrieve accurate diagrams. With the diagonal loss, diagrams produced without using the existence probability remain close to the ground-truth one (\cref{tab:add-filtr-ablation-loss}). 

\end{document}